\documentclass[final]{cvpr}

\usepackage{times}
\usepackage{epsfig}
\usepackage{graphicx}
\usepackage{amsmath}
\usepackage{amssymb}
\usepackage{algorithm}
\usepackage{algorithmicx}
\usepackage{algpseudocode}
\usepackage{subfigure}
\usepackage{multirow}
\usepackage{diagbox}
\usepackage{amsthm}
\usepackage{bm}
\usepackage{booktabs}  

\renewcommand{\mathbf}{\boldsymbol}

\DeclareMathOperator*{\argmin}{arg\,min}

\usepackage{dsfont}

\usepackage[pagebackref=true,breaklinks=true,colorlinks,bookmarks=false]{hyperref}

\def\cvprPaperID{6666} 
\def\confYear{CVPR 2021}

\begin{document}
	\newcolumntype{L}[1]{>{\raggedright\arraybackslash}p{#1}}
	\newcolumntype{C}[1]{>{\centering\arraybackslash}p{#1}}
	\newcolumntype{R}[1]{>{\raggedleft\arraybackslash}p{#1}}
	\title{Masked Surfel Prediction \\ for Self-Supervised Point Cloud Learning}
	
	\author{{Yabin Zhang$^1$ \quad Jiehong Lin$^2$ \quad Chenhang He$^1$ \quad Yongwei Chen$^2$ \quad Kui Jia$^2$ \quad Lei Zhang$^1$} \\
		$^1$Hong Kong Polytechnic University  \qquad $^2$South China University of Technology\\
		{\tt\small csybzhang@comp.polyu.edu.hk}
	}
	
	\maketitle
	
	\begin{abstract}
		Masked auto-encoding is a popular and effective self-supervised learning approach to point cloud learning. However, most of the existing methods reconstruct only the masked points and overlook the local geometry information, which is also important to understand the point cloud data. 
		In this work, we make the first attempt, to the best of our knowledge, to consider the local geometry information explicitly into the masked auto-encoding, and propose a novel Masked Surfel Prediction (MaskSurf) method. Specifically, given the input point cloud masked at a high ratio, we learn a transformer-based encoder-decoder network to estimate the underlying masked surfels by simultaneously predicting the surfel positions (i.e., points) and per-surfel orientations (i.e., normals). The predictions of points and normals are supervised by the Chamfer Distance and a newly introduced Position-Indexed Normal Distance in a set-to-set manner. Our MaskSurf is validated on six downstream tasks under three fine-tuning strategies. In particular, MaskSurf outperforms its closest competitor, Point-MAE, by 1.2\% on the real-world dataset of ScanObjectNN under the OBJ-BG setting, justifying the advantages of masked surfel prediction over masked point cloud reconstruction. Codes will be available at \url{https://github.com/YBZh/MaskSurf}. 
	\end{abstract}
	
	\section{Introduction} \label{Sec:intro}
	
	While deep learning has achieved great successes on various computer vision tasks, \eg, image classification \cite{krizhevsky2012imagenet,he2016deep}, object detection \cite{girshick2015fast,tian2019fcos}, segmentation \cite{ronneberger2015u,he2017mask}, image restoration \cite{dong2015image,zhang2017beyond}, as well as point cloud understanding \cite{qi2017pointnet,qi2017pointnet++}, training deep models usually requires a large amount of labeled data with human annotations, which are expensive in practice.  
	To solve this issue, self-supervised learning (SSL) \cite{chen2020improved,devlin2018bert,yu2021point} has been proposed to learn effective feature representations from unlabeled data. Generally speaking, SSL generates supervision signals from the data themselves by adopting various pretext tasks, such as contrastive learning \cite{he2020momentum,chen2020simple}, masked auto-encoding \cite{devlin2018bert,he2021masked,yu2021point}, rotation estimation \cite{gidaris2018unsupervised,poursaeed2020self},  jigsaw puzzles \cite{noroozi2016unsupervised} and so on \cite{afham2022crosspoint,grill2020bootstrap}.
	
	\begin{figure}[tb]
		\begin{center}
			\includegraphics[width=0.99\linewidth]{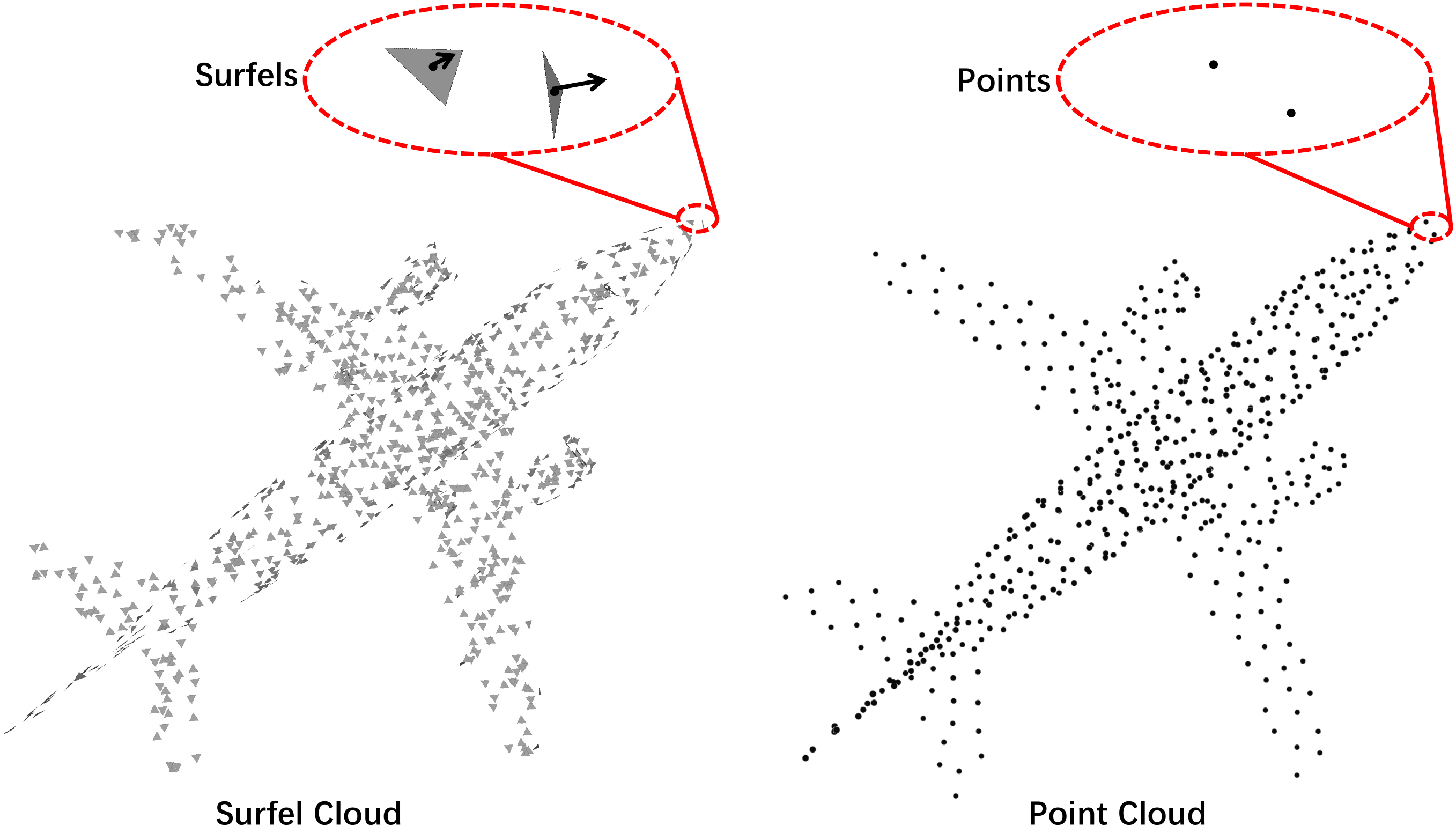}
		\end{center}
		\vspace{-0.2cm}
		\caption{Illustrations of the surfel cloud and point cloud, where surfels can capture more local geometry information than points.
		} \label{Fig:cloud_comparison}
		\vspace{-0.3cm}
	\end{figure}

	Among those pretext tasks, masked auto-encoding has demonstrated its effectiveness in many applications \cite{devlin2018bert,he2021masked,wei2021masked,tong2022videomae,yu2021point,pang2022masked}, including point cloud learning \cite{yu2021point,pang2022masked}. Specifically, by masking a portion of input data (\eg, points in point cloud processing), an auto-encoder is learned to reconstruct the masked data from the unmasked data. 
	In this manner, the encoder is expected to learn semantic feature representations, which could be readily applied to various downstream tasks. The popular masked auto-encoding based point cloud learning methods usually adopt different masking strategies and backbones, but they all reconstruct the masked points as the pretext task \cite{wang2021unsupervised,yu2021point,pang2022masked}.

	Though masked auto-encoding has achieved impressive progresses in self-supervised point cloud learning \cite{wang2021unsupervised,yu2021point,pang2022masked}, reconstructing the masked points only may sacrifice the local geometry information of point cloud. Though local geometry could be estimated from the point cloud data \cite{tatarchenko2018tangent,bae2008method,ran2022surface}, existing point cloud models \cite{qi2017pointnet,qi2017pointnet++,wang2019dynamic} are not effective to learn such local geometry. This can be validated by the fact that enhancing the point cloud inputs with local geometry (\eg, normal) could significantly boost the performance of point cloud models \cite{qi2017pointnet++,ran2022surface}, demonstrating the complementarity between the point location and local geometry in point cloud representation.

	With the above consideration, we propose to incorporate local geometry into the masked auto-encoding explicitly for more effective point cloud understanding.
	Specifically, we make the first attempt to employ the surface element, \ie, surfel \cite{pfister2000surfels}, for point cloud learning. 
	The vanilla surfel is originally introduced for 3D rendering, and it comprises both shape (\ie, surfel position and orientation) and shade (\ie, multiple levels of texture colors) data \cite{pfister2000surfels}.  The surface geometry is mainly described by its shape, while the shade information is more relevant to view synthesis and rendering. Considering that the goals of point cloud understanding are different from 3D rendering, we adopt a simplified surfel representation with only shape data of 3D position and orientation.
	As shown in Fig. \ref{Fig:cloud_comparison}, even the simplified surfel representation can capture more local geometry information of the surface over raw points. With surfel as the modeling element, different from those works predicting the point cloud \cite{wang2021unsupervised,yu2021point,pang2022masked}, we propose a \textbf{Mask}ed \textbf{Surf}el Prediction (MaskSurf) network to predict the underlying surfel cloud from the masked point cloud.  
	
	Following \cite{yu2021point,pang2022masked}, we first group the point cloud into several local patches and randomly mask a large portion of them. As illustrated in Fig. \ref{Fig:method},
	instead of reconstructing the masked point patches from unmasked point patches \cite{pang2022masked}, we predict the masked surfels \cite{pfister2000surfels} by simultaneously estimating the surfel positions (\ie, points) and per-surfel orientations (\ie, normals) in a set-to-set manner. 
	The point estimation is supervised by the Chamfer Distance (CD) \cite{fan2017point}, while a novel Position-Indexed Normal Distance (PIND) is proposed for point-paired normal prediction. 
	As analyzed in Sec. \ref{Subsec:analyses}, with surfel prediction, the learned features could capture more geometry information compared to the point only reconstruction \cite{pang2022masked}.

	Given the pre-trained encoder with MaskSurf, we validate its effectiveness on six downstream tasks, including object classification on real-world and synthetic datasets, few-shot learning, domain generalization, part segmentation and semantic segmentation.
	For each downstream task, we adopt various fine-tuning strategies \cite{he2020momentum,he2021masked}, including transferring features protocol, linear classification protocol and non-linear classification protocol.
	Our MaskSurf outperforms its closest competitor \cite{pang2022masked} on all downstream tasks under all strategies, justifying the advantage of masked surfel prediction over masked point cloud reconstruction. Notably, MaskSurf achieves $91.22\%$ accuracy on the real-world dataset of ScanObjectNN in the OBJ-BG setting, boosting  Point-MAE \cite{pang2022masked} by $1.2\%$.

	\section{Related Work}
	
	\subsection{Self-supervised Learning for Point Cloud}
	
	SSL aims to learn efficient feature representation from unlabeled training samples using self-generated supervision signals \cite{he2021masked,he2020momentum,chen2020improved,chen2020simple,grill2020bootstrap,devlin2018bert,yu2021point,pang2022masked}.  It is particularly important for 3D point cloud analysis, since the collection and annotation of point cloud data are much more expensive than 2D images. Popular SSL methods for point cloud include reconstruction \cite{yang2018foldingnet,gadelha2018multiresolution,zhao20193d,wang2021unsupervised,chen2021shape,han2019multi,zhou2022self,yu2021point,liu2022masked,pang2022masked,zhang2022point,xu2022cp,fu2022pos}, instance contrastive feature learning \cite{rao2020global,sanghi2020info3d}, consistency feature learning against augmentations \cite{huang2021spatio}, and other pretext tasks \cite{sauder2019self,poursaeed2020self,afham2022crosspoint}. 
	Among these methods, the masked auto-encoding \cite{wang2021unsupervised,zhou2022self,yu2021point,pang2022masked} has been receiving more and more attention recently. 
	
	Specifically, given an input point cloud masked at a high ratio, an encoder-decoder model is learned to reconstruct the masked points from the unmasked ones. In this way, the encoder could learn semantic feature representations, which can be readily applied to downstream tasks. However, the local geometry information may be overlooked by reconstructing the masked points only, since the local geometry is complementary to raw points for point cloud understanding \cite{qi2017pointnet++,ran2022surface}. To address this issue, we propose to explicitly incorporate the local geometry into the masked auto-encoding and develop a novel MaskSurf framework. In MaskSurf, we predict the underlying masked surfels by simultaneously estimating the surfel positions and per-surfel normals, resulting in more effective feature representations.
	
	\subsection{Local Geometry and Surfel Representation}
	The importance of local geometry in point cloud understanding has been widely acknowledged in the community \cite{alexa2003computing,pauly2003shape}, while normal is one of the most basic elements to represent local geometry information. Researchers typically enhance the point cloud data with point-wise normal for performance-boosting \cite{qi2017pointnet++,ran2022surface}.  What's more, given points as input, point-wise normal estimation is widely adopted as a regularization method to train the model \cite{lulu2020improving,rao2020global,xu2022cp}. 
	
	Surfel, \ie, surface element, is originally introduced as a rendering primitive, which provides a mere discretization of the geometry \cite{pfister2000surfels}.  Then, surfel has been widely adopted in surface reconstruction \cite{habbecke2007surface,weise2009hand} due to its conceptual simplicity.  
	The vanilla surfel comprises both shape and shade values, where the shape data describe the surface geometry, while the shade data are more relevant to rendering \cite{pfister2000surfels}.  	In this work, we adopt a simplified surfel representation with only shape data (\ie, 3D position and orientation) for model learning, considering the different objectives between point cloud understanding and 3D rendering.
	To our best knowledge, we are the first to apply surfel representation in self-supervised point cloud learning.

	
	\begin{figure*}[h]
		\begin{center}
			\includegraphics[width=0.96\linewidth]{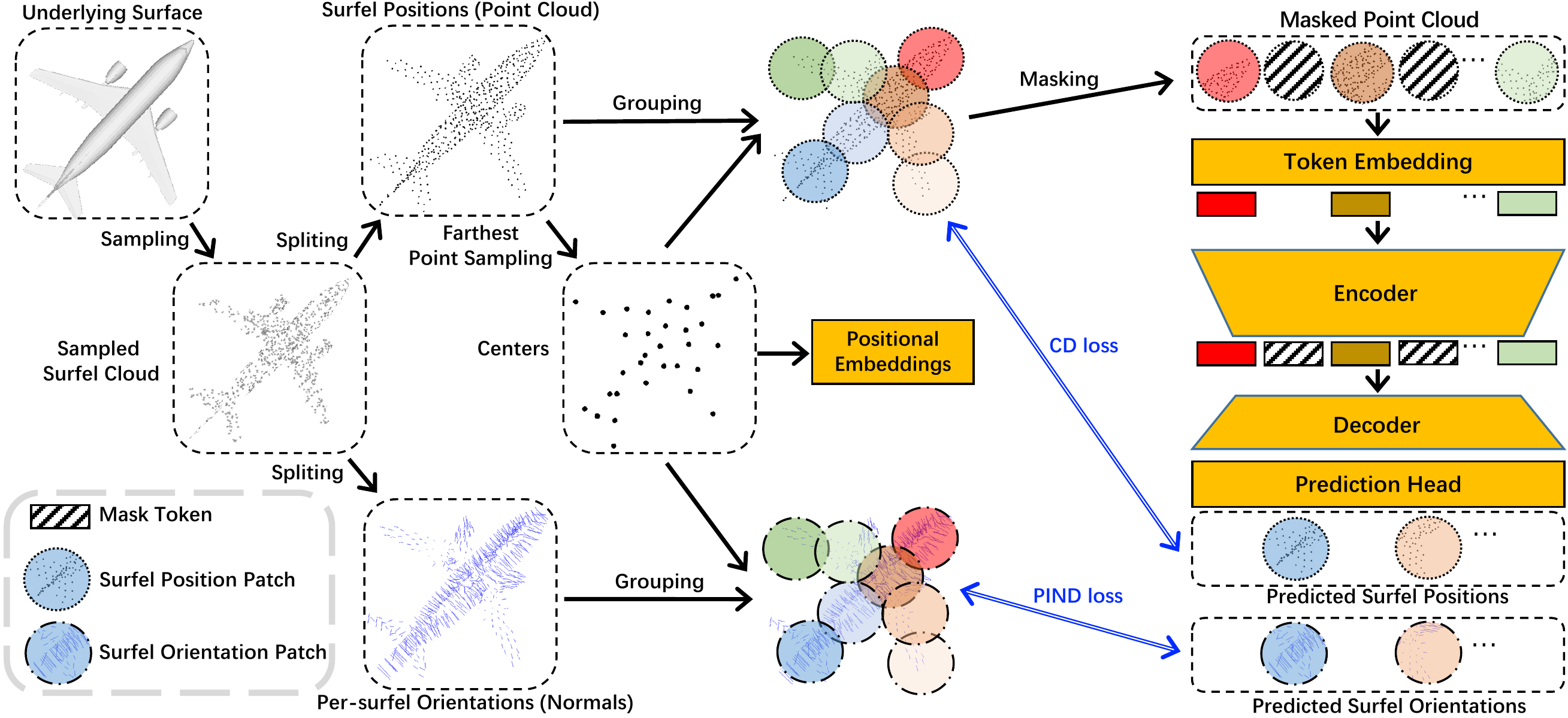}
		\end{center}
		\vspace{-0.2cm}
		\caption{The overall framework of our MaskSurf. We first sample a surfel cloud from a 3D surface and split the surfel cloud into surfel positions (\ie, points) and per-surfel orientations (\ie, normals), which are then grouped into patches. The point patches are randomly masked and embedded. Given embedded point patches, a transformer-based encoder-decoder network is learned to predict the underlying masked surfels by simultaneously predicting the surfel points and per-surfel normals, which are supervised by the Chamfer Distance (CD) and a newly introduced Position-Indexed Normal Distance (PIND), respectively.  
		} \label{Fig:method}
		\vspace{-0.2cm}
	\end{figure*}
	
	\section{Masked Surfel Prediction} \label{Sec:MSP}
	
	The overall framework of our MaskSurf is illustrated in Fig. \ref{Fig:method}. 
	Given masked and embedded point patches, we learn the transformer-based encoder and decoder to predict the underlying masked surfels by simultaneously predicting the surfel positions (\ie, points) and per-surfel orientations (\ie, normals). In the following subsections, we introduce the main components in detail.

	\subsection{Training Data Preparation}
	
	Considering that collecting high quality 3D samples in real world is expensive, most of the existing SSL methods \cite{wang2021unsupervised,yu2021point,pang2022masked} sample training data from synthetic 3D dataset (\eg, ShapeNet \cite{shapenet2015}). 
	Following this strategy, we sample a surfel cloud with $M$ surfels $\mathbf{S} \in \mathbb{R}^{M\times 6}$ from a synthetic 3D surface.  
	We then split the surfel cloud into surfel positions (\ie, points) $\mathbf{X} \in \mathbb{R}^{M\times 3}$ and per-surfel orientations (\ie, normals) $\mathbf{N} \in \mathbb{R}^{M\times 3}$. The masked point cloud will be used as the model input, while the normals will be only used to supervise the prediction of surfel orientations (see Sec. \ref{subsec:loss} for details). 
	
	We sample $N$ points from the point cloud $\mathbf{X}$ as patch centers $\mathbf{C} \in \mathbb{R}^{N\times 3}$ via the Farthest Point Sampling (FPS) method \cite{qi2017pointnet++}.  Then, for each center we introduce $N$ irregular point patches $\mathbf{P} \in \mathbb{R}^{N \times K \times 3}$ by selecting the $K$ nearest points around the center via the K-Nearest Neighborhood (KNN) method:
	\begin{equation} \label{Equ:knn}
	\mathbf{P} = KNN (\mathbf{X}, \mathbf{C}).
	\end{equation}
	Each point patch is then normalized by subtracting the center point from the point coordinates for better convergence. 
	Note that the point patches $\mathbf{P}$ may overlap if two patch centers in $\mathbf{C}$ are close to each other. 
	
	Following \cite{pang2022masked}, we mask each patch separately with a large ratio of point patches, keeping the information complete in each patch with rare patch overlap. More specifically, given a masking ratio $m\in (0,1)$, the masked point patches and unmasked point patches are denoted as $\mathbf{P}_{mask} \in \mathbb{R}^{mN \times K \times 3}$ and $\mathbf{P}_{vis} \in \mathbb{R}^{(1-m)N \times K \times 3}$, respectively. We then apply the same grouping (cf. Equ. (\ref{Equ:knn})) and masking strategies to the per-surfel normals $\mathbf{N}$, resulting in the masked normal patches $\mathbf{N}_{mask} \in \mathbb{R}^{mN \times K \times 3}$ and unmasked normal patches $\mathbf{N}_{vis} \in \mathbb{R}^{(1-m)N \times K \times 3}$.
	The masked patch centers $\mathbf{C}_{mask} \in \mathbb{R}^{mN\times 3}$ and unmasked patch centers $\mathbf{C}_{vis} \in \mathbb{R}^{(1-m)N\times 3}$ are similarly introduced for the usage of positional embedding. 
	
	The unmasked point patches $\mathbf{P}_{vis}$ are adopted as input to the following encoder model, while the masked point patches $\mathbf{P}_{mask}$ and masked normal patches $\mathbf{N}_{mask}$ are employed as the prediction supervision, which is detailed in the following subsections. 
	
	\subsection{Model Architecture}
	
	\noindent\textbf{Token Embedding.} 
	Before forwarding the visible point patches $\mathbf{P}_{vis}$ to the encoder, we first embed them via token embedding. Following \cite{pang2022masked}, we instantiate the token embedding with a lightweight PointNet \cite{qi2017pointnet}, which is composed of multi-layer perceptrons (MLP) and a max pooling layer. The embedded visible tokens $\mathbf{T}_{vis} \in \mathbb{R}^{(1-m)N \times D}$ are then induced as:
	\begin{equation}
	\mathbf{T}_{vis} = PointNet(\mathbf{P}_{vis}).
	\end{equation}
	
	\noindent\textbf{Encoder.} 
	We construct the encoder with standard Transformer blocks \cite{vaswani2017attention}. Only the visible tokens $\mathbf{T}_{vis}$ are encoded, while the masked patches are not exposed to the encoder. This is not only computationally efficient but also avoids early leakage of the position information of masked patches \cite{pang2022masked}. Considering that the point patches are represented with normalized coordinates, we add in each transformer block the path-wise Positional Embedding (PE) to provide patch location information. Following the common practice \cite{yu2021point,pang2022masked}, we adopt a learnable MLP as the PE, \ie, $PE_e$: $\mathbb{R}^{(1-m)N\times 3} \to \mathbb{R}^{(1-m)N\times D}$, which maps coordinates of the visible patch centers $\mathbf{C}_{vis}$ to the embedding dimension $D$. 
	Finally, the encoded visible tokens $\mathbf{T}_{e} \in \mathbb{R}^{(1-m)N \times D}$ are formulated as:
	\begin{equation}
	\mathbf{T}_{e} = Encoder(\mathbf{T}_{vis}, PE_e(\mathbf{C}_{vis})).
	\end{equation}
	
	\noindent\textbf{Decoder.} 
	Similar to the encoder, we also build the decoder with standard Transformer but with fewer blocks. 
	The decoder takes the encoded visible tokens $\mathbf{T}_{e}$, the learnable mask tokens $\mathbf{T}_{m} \in \mathbb{R}^{mN \times D}$, and their PEs as inputs, and outputs the decoded mask tokens $\mathbf{T}_{d} \in \mathbb{R}^{mN \times D}$:
	\begin{equation}
	\mathbf{T}_{d} = Decoder(\mathbf{T}_{e}, \mathbf{T}_{m}, PE_d(\mathbf{C})),
	\end{equation}
	where $\mathbf{T}_{m}$ is the duplication of a learnable and patch-shared mask token of $D$ dimension, and $PE_d(\mathbf{C})$ is the PE for all tokens (\ie, visible and mask tokens).
	As in \cite{pang2022masked}, we adopt two separate PEs for encoder and decoder, respectively.
	
	\noindent\textbf{Prediction Head.} 
	Existing methods typically introduce self-supervision by reconstructing masked points \cite{pang2022masked,yu2021point}. 
	Considering that surfels capture more local geometry information than points, we propose to estimate the masked surfels by predicting the surfel point positions and per-surfel normals. 
	Specifically, taking the decoded mask tokens $\mathbf{T}_{d}$ as inputs, the prediction head outputs patch-wise vectors, which are then reshaped and split into surfel position patches and per-surfel normal patches:
	\begin{align}
	\mathbf{\widehat{PN}} &= Reshape(FC(\mathbf{T}_{d})), \\
	\mathbf{\widehat{P}}, \mathbf{\widehat{N}} &= Split(\mathbf{\widehat{PN}}),
	\end{align}
	where $\mathbf{\widehat{PN}} \in \mathbb{R}^{mN \times K \times 6}$ is the concatenation of predicted masked surfel position patches  $\mathbf{\widehat{P}} \in \mathbb{R}^{mN \times K \times 3}$ and the per-surfel normal patches $\mathbf{\widehat{N}} \in \mathbb{R}^{mN \times K \times 3}$, and $FC(\cdot)$ indicates one fully connected (FC) layer.
	
	\subsection{Loss Functions} \label{subsec:loss}
	To measure the performance of masked surfel prediction, we measure the estimation of masked surfel positions and per-surfel orientations in a set-to-set manner.
	For the convenience of expression, in the following development we define the loss functions on one surfel position patch  $\mathbf{p} \in \mathbb{R}^{K \times 3}$ and its corresponding normal patch $\mathbf{n} \in \mathbb{R}^{K \times 3}$, which are sampled from $\mathbf{P}_{mask}$ and $\mathbf{N}_{mask}$, respectively; similarly, the predicted masked surfel position  patch and normal patch are denoted as $\mathbf{\widehat{p}} \in \mathbb{R}^{K \times 3}$ and $\mathbf{\widehat{n}} \in \mathbb{R}^{K \times 3}$, respectively. The final loss is calculated by averaging over all masked patches.
	
	Following 3D reconstruction methods \cite{fan2017point,pang2022masked}, we adopt the following Chamfer Distance (CD) loss to measure the divergence of point patches:
	\begin{align} \label{Equ:cd_loss}
	\mathcal{L}_p = \frac{1}{K} \sum_{k=1}^{K} \min_{k' \in [1,K]} \| \mathbf{p}_{k} - \mathbf{\widehat{p}} _{k'} \|_2^2   \nonumber \\
	+ \frac{1}{K} \sum_{k=1}^{K} \min_{k' \in [1,K]} \| \mathbf{\widehat{p}} _{k}  - \mathbf{p}_{k'}  \|_2^2,
	\end{align}
	where $\mathbf{p}_{k} \in \mathbb{R}^3$ and $\mathbf{\widehat{p}} _{k} \in \mathbb{R}^3$ are the $k$-th row of $\mathbf{p}$ and $\mathbf{\widehat{p}}$, respectively.  The  $\mathbf{n}_{k}$ and $\mathbf{\widehat{n}}_{k}$ in the following Equ. (\ref{Equ:normal_loss}) are similarly defined.
	
	How to measure the prediction performance of position-paired normal patches in a set-to-set manner is less investigated. Here we propose the following Position-Indexed Normal Distance (PIND) loss to address this issue,:
	\begin{align} \label{Equ:normal_loss}
	\mathcal{L}_n = & \frac{1}{K} \sum_{k=1}^{K}  d\left( \mathbf{n}_{k}, \mathbf{\widehat{n}}_{\argmin_{k' \in [1,K]}  \| \mathbf{p}_{k} - \mathbf{\widehat{p}} _{k'} \|_2^2} \right)  \nonumber \\
	+&  \frac{1}{K} \sum_{k=1}^{K}  d\left(  \mathbf{\widehat{n}}_{k} ,\mathbf{n}_{\argmin_{k' \in [1,K]}  \| \mathbf{\widehat{p}} _{k}  - \mathbf{p}_{k'}  \|_2^2} \right),
	\end{align} 
	where $d(\mathbf{n}, \mathbf{\widehat{n}})$ is the absolute cosine angle distance between two normal vectors $\mathbf{n}, \mathbf{\widehat{n}} \in \mathbb{R}^3$:
	\begin{equation} \label{Equ:absolute_cosine}
	d(\mathbf{n}, \mathbf{\widehat{n}}) = 1 - \left| \frac{\mathbf{n}}{\lVert\mathbf{n}\rVert_2}  \frac{\mathbf{\widehat{n}}}{\lVert\mathbf{\widehat{n}}\rVert_2} \right|.
	\end{equation}
	Similar to the CD loss in Equ. (\ref{Equ:cd_loss}), for each normal in one set, we find its `nearest neighbor' in the other set and sum the distances up in the PIND loss.  However, there are two differences between CD and PIND losses. Firstly, in PIND, we find the nearest neighbor of each normal according to the distance between corresponding positions, instead of the distance between normals, because the normal must be paired with one position to represent the surfel. Secondly, we adopt the absolute value of the cosine distance, instead of the Euclidean distance in CD loss, because the unoriented normal is sufficient for the surfel prediction.  
	
	The overall loss function is therefore defined as:
	\begin{equation} \label{Equ:overall_loss}
	\mathcal{L}_{all} = \mathcal{L}_p + \alpha \mathcal{L}_n,
	\end{equation}
	where $\alpha$ is a hyper-parameter balancing the two terms. 
	

	\section{Experiments}
	
	Our model is pre-trained on the ShapeNet \cite{shapenet2015} dataset, and then it is validated on various downstream tasks, including object classification on real-world and synthetic datasets, few-shot learning, domain generalization, part segmentation and semantic segmentation. Finally, we make in-depth analyses of the proposed components.
	
	\subsection{Pre-training on ShapeNet} \label{Sec:exp_pretrain}
	We pre-train our model on the ShapeNet \cite{shapenet2015}, which includes about $51K$ single clean 3D meshes shared by $55$ categories. Following \cite{yu2021point,pang2022masked}, we split the vanilla dataset into a training subset and a test subset, and use only the training subset for pre-training. For each 3D mesh in the training subset, we sample $p=1,024$ surfels from the surface and then split them as surfel positions and per-surfel normals. Data augmentations of standard random scaling and translation are applied to the sampled points.	We set the point patch size $K=32$ and divide the $1,024$ points into $N=64$ point patches. 	We then randomly mask the point patches with  masking ratio of $m=0.6$ by default. The other masking strategies are analyzed in Sec. \ref{Subsec:analyses}.
	
	We construct the encoder with $12$ Transformer blocks, while the decoder is built with $4$ Transformer blocks, where each Transformer block has $384$ hidden dimensions and $6$ heads.
	The AdamW optimizer \cite{loshchilov2017decoupled} is adopted. The batch size is 128 and the weight decay is 0.05. The cosine learning rate schedule \cite{loshchilov2016sgdr} is adopted with the total training epochs of 300 and an initial learning rate of 0.001. In order to reconstruct the indexing points first, we linearly increase the $\alpha$ from 0 to 0.01 in the training process. The predicted surfel cloud is visualized in Fig. \ref{Fig:predicted_surfel_vis}. 
	
	\subsection{Fine-tuning on Downstream Tasks}
	
	On downstream tasks, we initialize the encoder with the pre-trained weight parameters, while the decoder part of MaskSurf is discarded.
	The following three strategies are adopted to fine-tune pre-trained models on downstream tasks:
	\begin{itemize}
		\item Transferring features protocol, where we fine-tune all weight parameters, including the pre-trained encoder and a randomly initialized non-linear classifier. 
		\item Linear classification protocol, where we freeze the pre-trained encoder and only fine-tune a randomly initialized linear classifier.
		\item Non-linear classification protocol, where we freeze the pre-trained encoder and only fine-tune a randomly initialized non-linear classifier.
	\end{itemize}
	In transferring features and non-linear classification protocols, we construct the non-linear classifier via three FC layers for all classification tasks following  \cite{pang2022masked}. 
	On segmentation tasks (\ie, part segmentation and semantic segmentation), we follow \cite{pang2022masked,yu2021point} to sample $2,048$ points and introduce $128$ point patches. We strictly follow \cite{pang2022masked} to construct the classifier for segmentation, which is detailed in the \textbf{supplementary material}.
	In the `Transformer' method, we train both the encoder and non-linear classifier from scratch, setting a fair baseline. 
	We adopt the standard voting strategy \cite{liu2019relation} in the testing stage on ModelNet40 dataset following \cite{pang2022masked}, while no voting is performed on the other datasets.
	Note that existing methods typically report the best result across multiple runs on the classification task; here, we advocate reporting more detailed results with standard deviation to reflect the performance fluctuation.

	\begin{table}[htp] \footnotesize
		\centering
		\caption{Classification results on the ScanObjectNN dataset. } \label{Tab:scanobjectnn}
		\begin{tabular}{L{26.9mm}C{13.5mm}C{13.5mm}C{13.5mm}}
			\hline
			Methods & OBJ-BG & OBJ-ONLY & PB-T50-RS  \\
			\hline
			PointNet \cite{qi2017pointnet} &73.3 & 79.2 & 68.0 \\
			SpiderCNN \cite{xu2018spidercnn} & 77.1 & 79.5 & 73.7 \\
			PointNet++ \cite{qi2017pointnet++} & 82.3 & 84.3 & 77.9 \\
			DGCNN \cite{wang2019dynamic}  & 82.8 & 86.2 & 78.1 \\
			PointCNN \cite{li2018pointcnn} & 86.1 & 85.5 & 78.5 \\
			BGA-DGCNN \cite{uy2019revisiting} & -- & -- & 79.7 \\
			GBNet \cite{qiu2021geometric} & -- & -- & 80.5 \\
			Simple View \cite{goyal2021revisiting} & -- & -- & 80.5$\pm$0.3 \\
			PRANet \cite{cheng2021net} & -- & -- & 81.0 \\
			PointMLP \cite{ma2022rethinking} & -- & -- & 85.4$\pm$0.3  \\
			Transformer \cite{vaswani2017attention}  & 79.86 & 80.55 & 77.24 \\
			\hline
			\hline
			\multicolumn{4}{c}{\textbf{Transferring features protocol}}\\
			\hline
			Transformer-OcCo \cite{yu2021point}  & 84.85 & 85.54 & 78.79 \\
			Point-BERT \cite{yu2021point} & 87.43 & 88.12 & 83.07 \\
			Point-MAE \cite{pang2022masked} & 90.02 & 88.29 & 85.18  \\
			MaskSurf (Ours) & \textbf{91.22} & \textbf{89.17} & \textbf{85.81}  \\
			\hline
			\multicolumn{4}{c}{Detailed results with standard deviation }\\
			\hline
			Point-MAE  \cite{pang2022masked} & 89.26$\pm$0.39  & 88.19$\pm$0.32 & 84.66$\pm$0.40 \\
			MaskSurf (Ours) &  \textbf{90.76}$\pm$0.53 & \textbf{88.74}$\pm$0.23 & \textbf{85.35}$\pm$0.24  \\
			\hline
			\hline
			\multicolumn{4}{c}{\textbf{Linear classification protocol}}\\
			\hline
			Point-MAE \cite{pang2022masked} & 81.07$\pm$0.00 & 82.10$\pm$0.00 & 71.48$\pm$0.00  \\
			MaskSurf (Ours) & \textbf{82.07}$\pm$0.00 & \textbf{83.48}$\pm$0.00 & \textbf{72.59}$\pm$0.00  \\
			\hline
			\hline
			\multicolumn{4}{c}{\textbf{Non-linear classification protocol}}\\
			\hline
			Point-MAE \cite{pang2022masked} & 82.56$\pm$0.22 & 86.29$\pm$0.08 & 75.64$\pm$0.12  \\
			MaskSurf (Ours) & \textbf{84.45}$\pm$0.21 & \textbf{86.45}$\pm$0.08 & \textbf{76.48}$\pm$0.09  \\
			\hline
		\end{tabular}
	\end{table}
	
	\vspace{0.1cm}
	\noindent\textbf{Object Classification on Real-World Dataset.}
	Compared to 2D images, collecting and annotating 3D objects in the real world are much more expensive.
	Considering that many synthetic 3D objects are available on the web \cite{shapenet2015,wu20153d}, there is a massive demand to facilitate the real-world 3D tasks using synthetic 3D data. 
	Therefore, we first validate our pre-trained models on the real-world dataset of ScanObjectNN \cite{uy2019revisiting}, which includes about $15K$ point cloud samples shared by $15$ categories. The objects are scanned indoor scene data, which are often cluttered with background and occluded by other objects. 
	
	We adopt three experiment variants: OBJ-BG, OBJ-ONLY and PB-T50-RS, which are detailed in the \textbf{supplementary material}. As illustrated in Tab. \ref{Tab:scanobjectnn}, our MaskSurf significantly boosts the vanilla Transformer baseline with absolute improvements of 11.36\%, 8.62\%, and 8.57\% on the settings of OBJ-BG, OBJ-ONLY, and PB-T50-RS, respectively. Meanwhile, MaskSurf consistently outperforms its closest SSL competitor Point-MAE \cite{ pang2022masked}, which is based on masked point cloud reconstruction, under all the three fine-tuning protocols, justifying the advantage of our masked surfel prediction. 
	
	\begin{table}[htp]
		\caption{Classification results on ModelNet40 dataset. `ST' indicates whether the backbone is a standard Transformer without any special design or inductive bias. `Our rep.' means that the result is reproduced or produced by us using the official codes. Note that Point-MAE \cite{ pang2022masked} only reports the result under the transferring features protocol in the original paper.} \label{Tab:modelnet40}
		\centering
		\begin{tabular}{l|cc}
			\hline
			Methods & ST? & Accuracy (\%) \\
			\hline
			PointNet \cite{qi2017pointnet} & -- & 89.2 \\
			PointNet++ \cite{qi2017pointnet++} & -- & 90.7 \\
			PointCNN \cite{li2018pointcnn} & -- & 92.5 \\
			KPConv \cite{thomas2019kpconv} & -- & 92.9 \\
			DGCNN \cite{wang2019dynamic} & -- & 92.9 \\
			RS-CNN \cite{liu2019relation} & -- & 92.9 \\
			PCT \cite{guo2021pct} & N & 93.2 \\
			PVT \cite{zhang2021pvt} & N & 93.6 \\
			PointTransformer \cite{zhao2021point} & N & 93.7 \\
			Transformer \cite{vaswani2017attention}  & Y & 91.4 \\
			\hline
			\hline
			\multicolumn{3}{c}{\textbf{Transferring features protocol}}\\
			\hline
			DGCNN + OcCo \cite{wang2021unsupervised} & -- & 93.0 \\
			DGCNN + STRL \cite{huang2021spatio} & -- & 93.1 \\
			DGCNN + FoldingNet \cite{yang2018foldingnet} & -- & 93.1 \\
			\hline
			Transformer-OcCo \cite{yu2021point}  & Y & 92.1 \\
			Point-BERT \cite{yu2021point} & Y & 93.2 \\
			Point-MAE \cite{pang2022masked} & Y & \textbf{93.8} \\
			Point-MAE (Our rep.) & Y & 93.27 \\
			MaskSurf (Ours) & Y & 93.40 \\
			\hline
			\multicolumn{3}{c}{Detailed results with standard deviation. }\\
			\hline
			Point-MAE (Our rep.) & Y &   93.06$\pm$0.18  \\
			MaskSurf (Ours) & Y &   \textbf{93.18}$\pm$0.15  \\
			\hline
			\hline
			\multicolumn{3}{c}{\textbf{Linear classification protocol}}\\
			\hline
			DGCNN + Multi-Task \cite{hassani2019unsupervised} & -- & 89.1 \\
			DGCNN + Self-Contrast \cite{du2021self} & -- & 89.6\\
			DGCNN + Jigsaw \cite{sauder2019self} & -- & 90.6 \\
			DGCNN + FoldingNet \cite{yang2018foldingnet} & -- & 90.1 \\
			DGCNN + Rotation \cite{poursaeed2020self} & -- & 90.8 \\
			DGCNN + STRL \cite{huang2021spatio} & -- & 90.9 \\
			DGCNN + OcCo \cite{wang2021unsupervised} & -- & 89.2 \\
			DGCNN + CrossPoint \cite{afham2022crosspoint} & -- & 91.2 \\
			DGCNN + IAE \cite{yan2022implicit} & -- & 92.1 \\
			\hline
			Point-MAE (Our rep.) & Y &   91.41$\pm$0.00  \\
			MaskSurf (Ours) & Y &   \textbf{92.26}$\pm$0.00  \\
			\hline
			\hline
			\multicolumn{3}{c}{\textbf{Non-linear classification protocol}}\\
			\hline
			Point-MAE (Our rep.)          & Y &   92.59$\pm$0.13  \\
			MaskSurf (Ours) & Y &   \textbf{93.44}$\pm$0.03  \\
			\hline
		\end{tabular}
	\end{table}
	
	\vspace{0.1cm}
	\noindent\textbf{Object Classification on Synthetic Datasets.}
	Besides the real-world dataset discussed above, we also test MaskSurf on synthetic datasets of ModelNet40 \cite{wu20153d} and ShapeNet \cite{shapenet2015}. 
	Compared to the real-world ScanObjectNN dataset, these two tasks are much easier since the input point clouds are clean and complete, resulting in a smaller gap to the dataset used for pre-training. 
	Note that the ShapeNet dataset is also used in the pre-training stage, as detailed in Sec. \ref{Sec:exp_pretrain}.
	The ModelNet40 includes $12,311$ clean 3D CAD models for $40$ categories. Following the standard split, $9843$ and $2468$ samples are used for training and testing, respectively.

	\begin{table}[htp]
		\caption{Classification results on the ShapeNet dataset. } \label{Tab:shapenet}
		\centering
		\begin{tabular}{l|c}
			\hline
			Methods  & Accuracy (\%) \\
			\hline
			Transformer \cite{vaswani2017attention}  & 90.86$\pm$0.05 \\
			\hline
			\hline
			\multicolumn{2}{c}{\textbf{Transferring features protocol}}\\
			\hline
			Point-MAE  \cite{pang2022masked}           & \textbf{90.84}$\pm$0.02   \\
			MaskSurf (Ours)  & \textbf{90.84}$\pm$0.04    \\
			\hline
			\hline
			\multicolumn{2}{c}{\textbf{Linear classification protocol}}\\
			\hline
			Point-MAE  \cite{pang2022masked}          &  89.08$\pm$0.12   \\
			MaskSurf (Ours)  &  \textbf{89.62}$\pm$0.12   \\
			\hline
			\hline
			\multicolumn{2}{c}{\textbf{Non-linear classification protocol}}\\
			\hline
			Point-MAE   \cite{pang2022masked}         &  90.40$\pm$0.06   \\
			MaskSurf (Ours)  &  \textbf{91.09}$\pm$0.05   \\
			\hline
		\end{tabular}
	\end{table}

	Results on ModelNet40 and ShapeNet datasets are illustrated in Tab. \ref{Tab:modelnet40} and Tab. \ref{Tab:shapenet}, respectively.  Our MaskSurf consistently improves over PointMAE \cite{pang2022masked}, which is based on masked point cloud reconstruction, under all the three fine-tuning protocols.
	Specifically, under the transferring features protocol, different reconstruction-based SSL methods achieve comparable performance, since the two datasets are relatively easy.  
	Under more challenging settings (\ie, linear classification and non-linear classification protocols), where the pre-trained encoder is frozen, our MaskSurf shows more significant advantages over its closest competitor Point-MAE (\eg, 0.85\% on ModelNet40 and 0.69\% on ShapeNet under the non-linear classification protocol).  Note that such improvements are significant since the results are getting saturated on these two tasks.

	In addition, we have three interesting observations. Firstly, on the challenging real-world dataset of ScanObjectNN, the transferring features protocol is preferred, since there is a large domain gap between the synthetic pre-training data and the real-world testing data. The results of different methods vary on easier downstream tasks with synthetic samples. Specifically, under the transferring features protocol, Point-MAE achieves better results, while under the non-linear classification protocol, models pre-trained with MaskSurf are preferred.	This may be because fine-tuning the pre-trained encoder may degrade the local geometry perception ability of our MaskSurf. Secondly, on the ShapeNet dataset, under the non-linear classification protocol, only our MaskSurf  outperforms the fully-supervised Transformer baseline, justifying the advantages of the local geometry perception. 
	Finally, though the transformer backbone adopted in our MaskSurf is weaker than the DGCNN backbone used in most SSL methods, as presented in Tab. \ref{Tab:modelnet40}, MaskSurf still achieves better results than DGCNN-based SSL competitors on the ModelNet40 dataset, demonstrating its effectiveness. 
	

	\begin{table}[htb]
		\caption{Cross-domain generalization performance. `S' denotes the real-world ScanNet-10 dataset.} \label{Tab:domain_generalization}
		\centering
		\begin{tabular}{lcc}
			\hline
			Methods & ModelNet-10$\to$S & ShapeNet-10$\to$S\\
			\hline
			DGCNN \cite{wang2019dynamic} & 43.8$\pm$2.3 & 42.5$\pm$1.4 \\
			DANN \cite{ganin2016domain} & 42.1$\pm$0.6 & 50.9$\pm$1.0  \\
			PointDAN \cite{qin2019pointdan} & 44.8$\pm$1.4 & 45.7$\pm$0.7   \\		
			Transformer \cite{vaswani2017attention} & 44.43$\pm$2.38 & 42.62$\pm$1.45   \\
			\hline
			\hline
			\multicolumn{3}{c}{\textbf{Transferring features protocol}}\\
			\hline
			Point-MAE \cite{pang2022masked} & 47.16$\pm$1.51 & 46.67$\pm$0.03    \\
			MaskSurf (Ours) & \textbf{47.20}$\pm$0.95 & \textbf{48.26}$\pm$1.80  \\
			\hline
			\hline
			\multicolumn{3}{c}{\textbf{Linear classification protocol}}\\
			\hline
			Point-MAE \cite{pang2022masked} & 46.73$\pm$3.01  & 47.88$\pm$0.58 \\
			MaskSurf (Ours) & \textbf{46.90}$\pm$3.12 & \textbf{48.69}$\pm$1.19 \\
			\hline
			\hline
			\multicolumn{3}{c}{\textbf{Non-linear classification protocol}}\\
			\hline
			Point-MAE \cite{pang2022masked} & 40.31$\pm$0.02 & 40.93$\pm$0.03 \\
			MaskSurf (Ours) &  \textbf{46.13}$\pm$0.01 & \textbf{47.37}$\pm$0.02   \\
			\hline
		\end{tabular}
	\end{table}
	
	\vspace{0.1cm}
	\noindent\textbf{Domain Generalization.} 
	Applying models trained on synthetic domains to real-world applications has great practical value. We evaluate the cross-domain generalization performance of MaskSurf on the PointDA-10 dataset \cite{qin2019pointdan}, whose detailed information can be found in the \textbf{supplementary material}. Specifically, we adopt the synthetic 3D datasets of ModelNet-10 and ShapeNet-10 as the training set, and test the domain generalization performance on the real-world ScanNet-10 dataset with the model selection of training-domain validation \cite{gulrajani2020search}. As shown in Tab. \ref{Tab:domain_generalization}, our MaskSurf consistently outperforms its competitors, including the Transformer baseline and Point-MAE \cite{pang2022masked}.  

	\vspace{0.1cm}
	\noindent\textbf{Few-shot Learning.}
	We conduct the experiments of few-shot learning on the ScanObjectNN dataset under the ``$n$-way, $m$-shot'' setting, where $n$ is the number of randomly sampled classes and $m$ is the number of samples in each class.  The $n\times m$ samples are adopted for training, while we randomly sample $20$ unseen objects from each class for testing.
	We report the results of each setting with $10$ independent experiments.
	Results with $n=\{5,10\}$ and $m=\{10,20\}$ are presented in Tab. \ref{Tab:few_shot_scanobjectnn}. MaskSurf consistently outperforms its competitors under all fine-tuning protocols. Similar results can be observed on the ModelNet40 dataset. Please see the \textbf{supplementary material} for details.

	\begin{table}[htb] \small
		\caption{Few-shot classification performance on ScanObjectNN.} \label{Tab:few_shot_scanobjectnn}
		\centering
		\begin{tabular}{L{23.6mm}C{10.0mm}C{10.0mm}C{10.0mm}C{10.0mm}}
			\hline
			& \multicolumn{2}{c}{\textbf{5-way}} & \multicolumn{2}{c}{\textbf{10-way}} \\
			\cmidrule(r){2-3} 	\cmidrule(r){4-5}
			& 10-shot & 20-shot & 10-shot & 20-shot \\
			\hline
			Transformer \cite{vaswani2017attention} & 51.9$\pm$8.3 & 61.6$\pm$8.5 & 38.5$\pm$5.9 & 45.5$\pm$3.9 \\
			\hline
			\hline
			\multicolumn{5}{c}{\textbf{Transferring features protocol}}\\
			\hline
			Point-MAE \cite{pang2022masked} & 63.9$\pm$7.0 & 77.0$\pm$5.2 & 53.6$\pm$5.4 & 61.6$\pm$2.7 \\
			MaskSurf (Ours) &  \textbf{65.3}$\pm$4.9 & \textbf{77.4}$\pm$5.2 & \textbf{53.8}$\pm$5.3 & \textbf{63.2}$\pm$2.7  \\
			\hline
			\hline
			\multicolumn{5}{c}{\textbf{Linear classification protocol}}\\
			\hline
			Point-MAE \cite{pang2022masked} & 48.3$\pm$7.8 & 56.0$\pm$11.2 & 39.2$\pm$10.1 & 59.0$\pm$3.3 \\
			MaskSurf (Ours) & \textbf{51.0}$\pm$8.2 & \textbf{59.8}$\pm$7.9 & \textbf{41.7}$\pm$9.2 & \textbf{61.0}$\pm$3.4 \\
			\hline
			\hline
			\multicolumn{5}{c}{\textbf{Non-linear classification protocol}}\\
			\hline
			Point-MAE \cite{pang2022masked} & 56.4$\pm$6.8 & 67.2$\pm$6.5 & 44.3$\pm$6.2 & 50.8$\pm$3.6 \\
			MaskSurf (Ours) &  \textbf{60.8}$\pm$6.6 & \textbf{68.3}$\pm$6.7 & \textbf{46.6}$\pm$6.4 & \textbf{54.9}$\pm$3.5  \\
			\hline
		\end{tabular}
	\end{table}

	\vspace{0.1cm}
	\noindent\textbf{Part Segmentation.}
	We conduct part segmentation on the ShapeNetPart dataset \cite{yi2016scalable}, which includes $16,881$ samples shared by $16$ categories. As illustrated in Tab. \ref{Tab:partseg}, MaskSurf outperforms the Transformer baseline, and achieves comparable results to the state-of-the-art methods under the transferring features protocol.
	Note that neither Point-MAE nor our MaskSurf bring improvements to the Transformer baseline under the less-studied non-linear classification protocol, demonstrating the gap between reconstruction and segmentation tasks. Similar results can be observed in the following semantic segmentation task.
	
	
	\begin{table} [htb]
		\centering
		\caption{Part segmentation results on the ShapeNetPart dataset. The mean IoU across all categories, \ie, mIoU$_{c}$ (\%), and the mean IoU across all instances, \ie, mIoU$_{I}$ (\%) are reported. The IoU (\%) for each category is detailed in the \textbf{supplementary material}. } \label{Tab:partseg}
		\begin{tabular}{lcc}
			\hline
			Methods & mIoU$_{c}$ & mIoU$_{I}$  \\
			\hline
			PointNet \cite{qi2017pointnet} & 80.39 & 83.7 \\
			PointNet++ \cite{qi2017pointnet++} & 81.85 & 85.1 \\
			DGCNN \cite{wang2019dynamic} & 82.33 & 85.2 \\
			Transformer \cite{vaswani2017attention} & 83.42 & 85.1  \\
			\hline
			\hline
			\multicolumn{3}{c}{\textbf{Transferring features protocol}}\\
			\hline
			Point-BERT \cite{yu2021point} & 84.11 & 85.6   \\
			Point-MAE \cite{pang2022masked} & 84.19 & \textbf{86.1}  \\
			MaskSurf (Ours) & \textbf{84.36} & \textbf{86.1}  \\
			\hline
			\hline
			\multicolumn{3}{c}{\textbf{Non-linear classification protocol}} \\
			\hline
			Point-MAE \cite{pang2022masked} & 83.13 & 84.6   \\
			MaskSurf (ours) & \textbf{83.30} & \textbf{85.3}  \\
			\hline
		\end{tabular}
	\end{table}
	
	\vspace{0.1cm}
	\noindent\textbf{Semantic Segmentation.}
	We conduct the semantic segmentation on the Stanford 3D Indoor Scene Dataset (S3DIS) \cite{armeni20163d}, which contains 6 large-scale indoor areas with points shared by 13 classes.
	Different from most segmentation methods \cite{qi2017pointnet,li2018pointcnn,thomas2019kpconv} that adopt both $xyz$ and $rgb$ colors as input, we adopt the $xyz$ as input since the pre-trained model only accepts point cloud data. However, as shown in Tab. \ref{Tab:semantic_seg}, MaskSurf still shows clear improvement over the competiting methods, validating its advantages in feature representation. 
	
	\begin{table}[htb]
		\caption{Semantic segmentation results on the S3DIS Area 5.} \label{Tab:semantic_seg}  
		\centering
		\begin{tabular}{lcccc}
			\hline
			Methods & Input & OA & mAcc & mIoU \\
			\hline
			PointNet \cite{qi2017pointnet} & $xyz$+$rgb$ & -- & 49.0 & 41.1 \\
			PointCNN \cite{li2018pointcnn} & $xyz$+$rgb$ & 85.9 & 63.9 & 57.3  \\
			KPConv \cite{thomas2019kpconv} & $xyz$+$rgb$ & -- & 72.8 & 67.1 \\
			Transformer \cite{vaswani2017attention} & $xyz$ & 86.8 & 68.6 & 60.0   \\
			\hline
			\hline
			\multicolumn{5}{c}{\textbf{Transferring features protocol}} \\
			\hline
			Point-MAE \cite{pang2022masked} &  $xyz$ & 87.4 & 69.4 & 61.0    \\
			MaskSurf (Ours) &  $xyz$ &  \textbf{88.3} & \textbf{69.9} & \textbf{61.6}   \\
			\hline
			\hline
			\multicolumn{5}{c}{\textbf{Non-linear classification protocol}}\\
			\hline
			Point-MAE \cite{pang2022masked} & $xyz$ & 85.3 & 65.4 & 56.1    \\
			MaskSurf (Ours) & $xyz$ & \textbf{86.2} & \textbf{66.6} & \textbf{56.6}    \\
			\hline
		\end{tabular}
	\end{table}
	
	\noindent\textbf{Summary on Downstream Tasks.} 
	Our MaskSurf demonstrates considerable advantages on more challenging tasks (\eg, the ScanObjectNN dataset and the linear classification protocol), while results of different SSL methods are comparable on easier tasks (\eg, classification on ModelNet40 and ShapeNet under the transferring features protocol). Moreover, the generation-based SSL methods (\eg, Point-BERT, Point-MAE, and our MaskSurf) bring marginal improvement in segmentation tasks, implying the need for segmentation-specific SSL strategies. 
	
	\subsection{Analyses and Discussions} \label{Subsec:analyses}
	
	
	\begin{figure}[t]
		\begin{center}
			\includegraphics[width=0.99\linewidth]{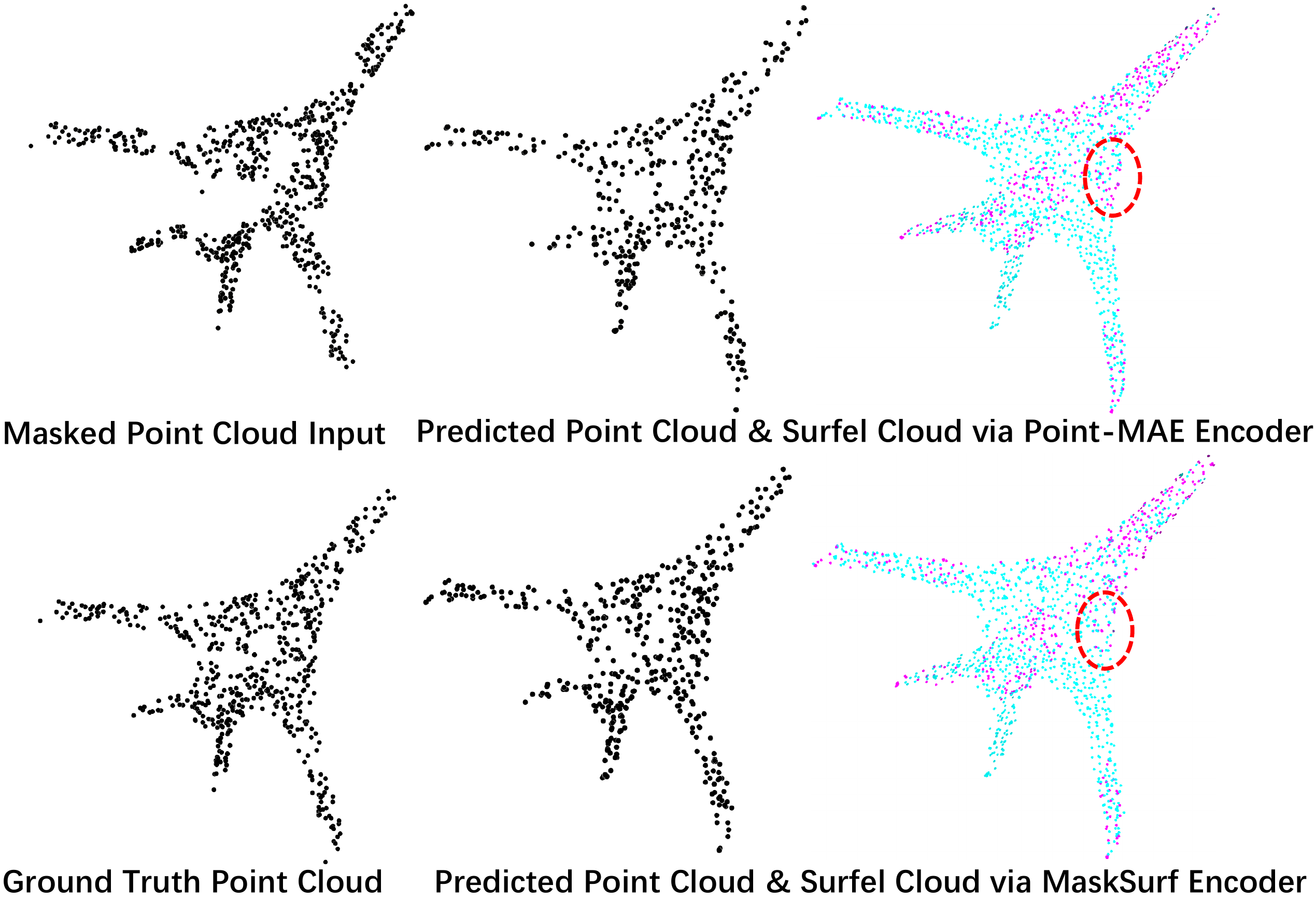}
		\end{center}
		\vspace{-0.2cm}
		\caption{Visualization of the predicted point cloud and surfel cloud with frozen encoders. In surfel cloud, the blue color means that the unoriented angular difference between estimated surfel normal and ground truth normal is less than 30 degrees, while the red color means that the unoriented angular difference is larger than 30 degrees.
		} \label{Fig:predicted_surfel_vis}
		\vspace{-0.2cm}
	\end{figure}
	
	\noindent\textbf{Pre-trained Encoders.}   We freeze the pre-trained encoders and learn decoders from scratch with our proposed surfel prediction objective (cf. Equ. (\ref{Equ:overall_loss})). 
	As shown in Tab. \ref{Tab:fix_encoder_analyses}, MaskSurf achieves better surfel prediction performance (\eg, lower $\mathcal{L}_p$  and $\mathcal{L}_n$) than Point-MAE, which is also visualized in Fig. \ref{Fig:predicted_surfel_vis}. 
	\begin{table}[htb] \small
		\centering
		\caption{Quantitive analyses of the surfel prediction on the ShapeNet test subset with frozen encoders.  The quality of point reconstruction and normal prediction are measured by values of  $\mathcal{L}_p$ and  $\mathcal{L}_n$, respectively.} \label{Tab:fix_encoder_analyses}
		\begin{tabular}{l|cc}
			\hline
			Methods & $\mathcal{L}_p$ $\downarrow$ &  $ \mathcal{L}_n$ $\downarrow$  \\
			\hline
			Poine-MAE \cite{pang2022masked} & 2.26 $\times$ 1e-3   & 0.57 \\
			MaskSurf (Ours)  & \textbf{2.19} $\times$ 1e-3   & \textbf{0.51} \\
			\hline
		\end{tabular}
	\end{table}

\begin{figure}[h]
		\vspace{-0.2cm}
		\begin{center}
			\includegraphics[width=0.82\linewidth]{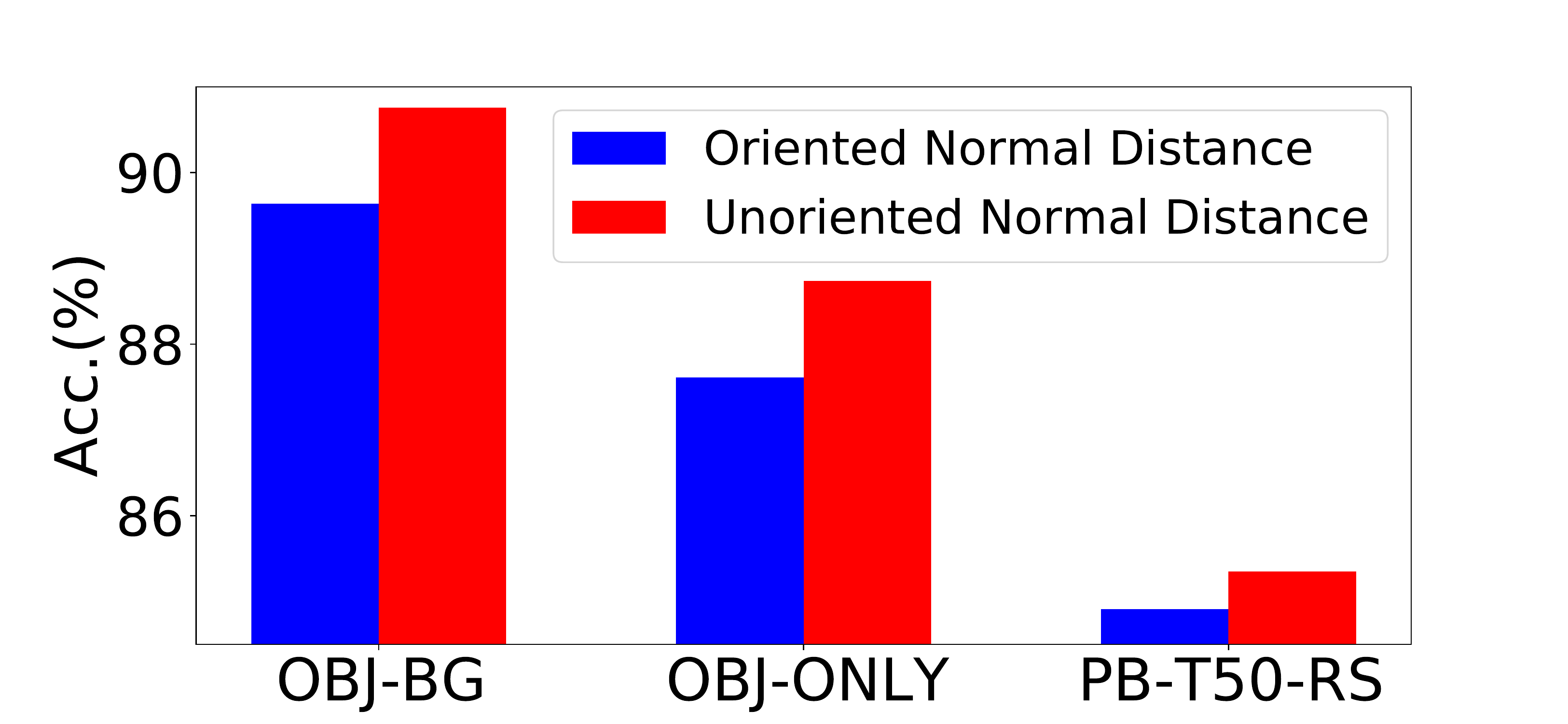}
		\end{center}
		\vspace{-0.2cm}
		\caption{Classification results on the PB-T50-RS setting of ScanObjectNN dataset with various normal distance.
		} \label{Fig:oriented_or_unoriented_normal}
	\end{figure}

	\noindent\textbf{Variants of Normal Distance.} Results with unoriented normal distance (\ie, Equ. (\ref{Equ:absolute_cosine})) and oriented normal distance (\ie, Equ. (\ref{Equ:absolute_cosine}) without absolute function) are compared in Fig. \ref{Fig:oriented_or_unoriented_normal}. Unoriented normal distance presents clear advantage, which is adopted as the default setting. 

\begin{figure}[h]
		\begin{center}
			\includegraphics[width=0.82\linewidth]{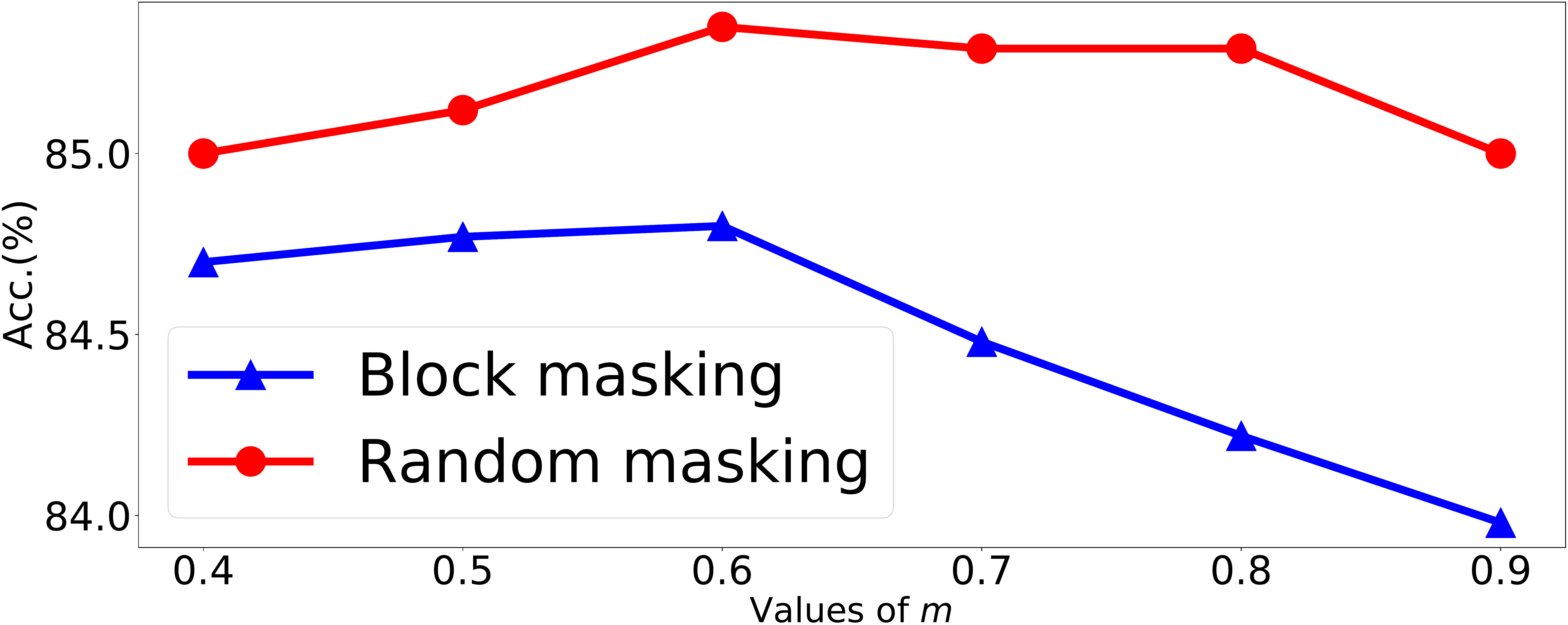}
		\end{center}
		\vspace{-0.2cm}
		\caption{Classification results on the PB-T50-RS setting of ScanObjectNN dataset with various masking strategies.
		} \label{Fig:masking_strategy}
	\end{figure}

	
	\noindent\textbf{Masking Strategies.} 
	As illustrated in Fig. \ref{Fig:masking_strategy}, random masking leads to higher accuracy over the block masking strategy \cite{yu2021point}, and the best results are achieved when the mask ratio $m=0.6$, which is adopted as the default setting.
	
\begin{figure}[h]
		\begin{center}
			\includegraphics[width=0.8\linewidth]{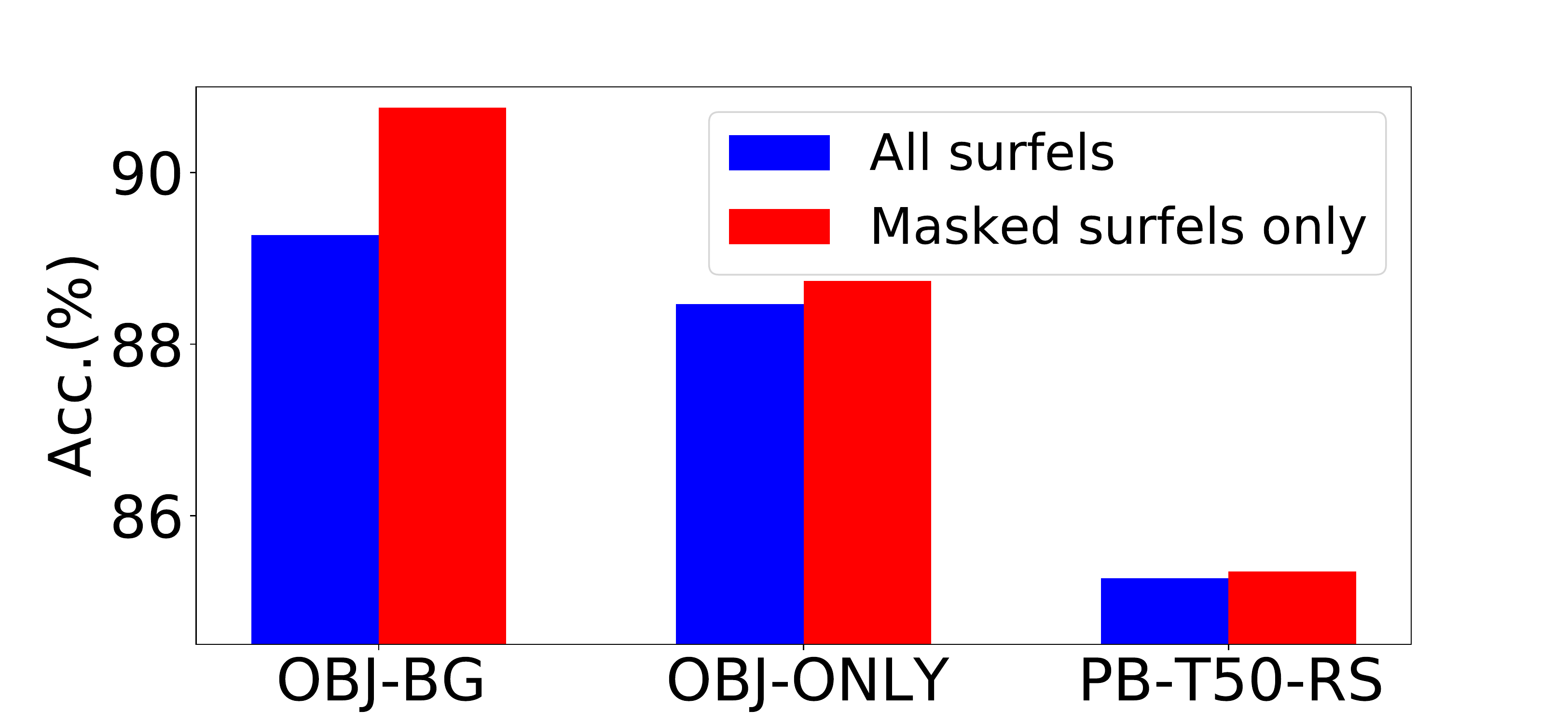}
		\end{center}
		\vspace{-0.2cm}
		\caption{Classification results on ScanObjectNN dataset with different reconstruction objectives.
		} \label{Fig:rec_all_or_rec_masked}
	\end{figure}

	
	\noindent\textbf{Reconstructing Masked Surfels or All Surfels?}  
	Similar to observations in \cite{he2021masked}, better results are achieved by reconstructing masked parts only, as shown in Fig. \ref{Fig:rec_all_or_rec_masked}. 

\begin{figure}[h]
		\begin{center}
			\includegraphics[width=0.8\linewidth]{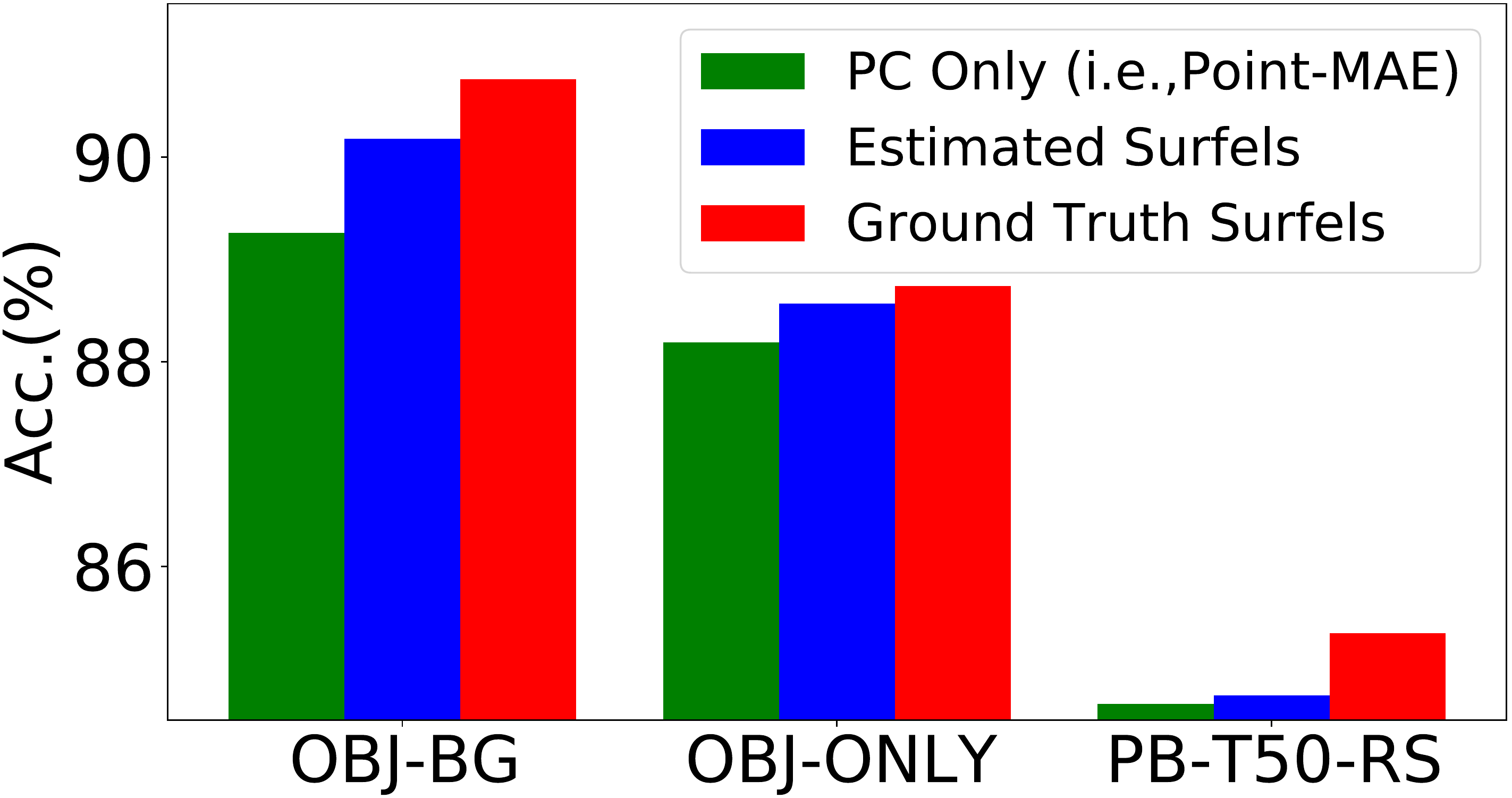}
		\end{center}
		\vspace{-0.2cm}
		\caption{Classification results with various reconstruction targets. `PC' is short for point cloud.
		} \label{Fig:estimated_normal}
	\end{figure}

	\noindent\textbf{Results with Estimated Surfels.} 
	To pre-train MaskSurf on a pure point cloud dataset (\eg, when the underlying 3D surfaces are not accessible), we could estimate the surfel cloud from the point cloud \cite{tatarchenko2018tangent}  and adopt the estimated surfels as the supervision. As illustrated in Fig. \ref{Fig:estimated_normal}, although estimated surfels result in lower performance than ground truth surfels, they still lead to better performance than reconstructing point cloud only (\ie, Point-MAE), revealing the broader applications of MaskSurf.

\begin{figure}[h]
		\begin{center}
			\includegraphics[width=0.82\linewidth]{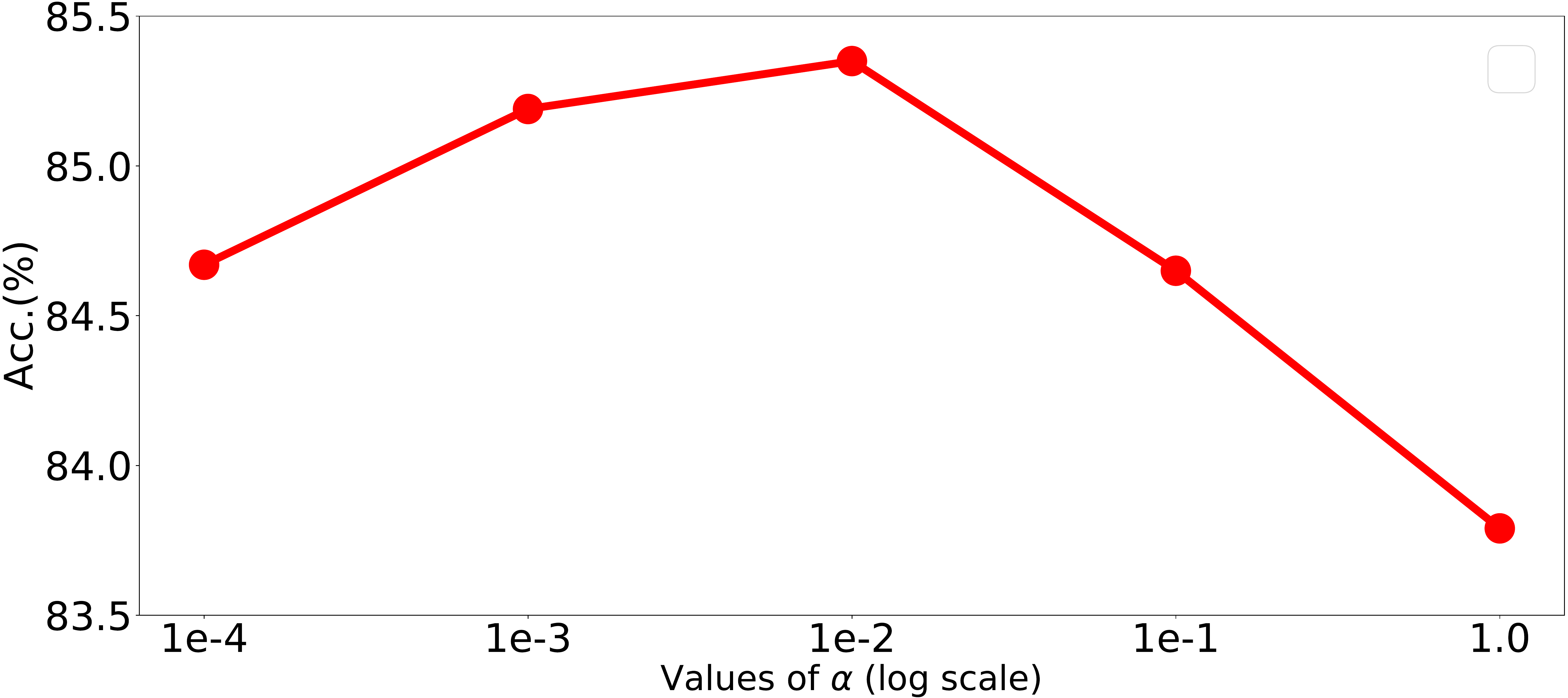}
		\end{center}
		\vspace{-0.2cm}
		\caption{Classification results on the PB-T50-RS setting of ScanObjectNN  dataset with different $\alpha$ values.
		} \label{Fig:alpha}
	\end{figure}

	\noindent\textbf{Hyper-parameter $\alpha$}. 
	As illustrated in Fig. \ref{Fig:alpha}, $\alpha=0.01$ leads to the best performance, which is adopted as the default setting in all experiments. 

	
	\noindent\textbf{Complexity Analysis.} 
	As illustrated in Tab. \ref{Tab:parameter_complexity}, MaskSurf introduces about 0.1\% additional parameters and multiply-accumulates (MACs) compared to Point-MAE in the pre-training stage, while it has the same complexity as the baseline Transformer on downstream tasks.

	\begin{table}[htb]\small
		\centering
		\caption{Illustrations of the model parameters and computational complexity. The `Fine-tuning' is reported on the downstream classification tasks.}
		\label{Tab:parameter_complexity}
		\begin{tabular}{lcccc}
			\hline
			\multirow{2}{*}{Methods} & \multicolumn{2}{c}{\textbf{Pre-training}} & \multicolumn{2}{c}{\textbf{Fine-tuning}} \\
			\cmidrule(r){2-3} 	\cmidrule(r){4-5}
			& Params & MACs & Params & MACs \\
			\hline
			Transformer \cite{vaswani2017attention}  &  -- & --& 22.1M & 2.4G \\
			Point-MAE \cite{pang2022masked}    & 29.0M & 2.5G & +0\% & +0\% \\
			MaskSurf (Ours)    & +0.127\% & +0.069\% & +0\% & +0\% \\
			\hline
		\end{tabular}
	\end{table}
	
	\section{Conclusion}
	
	We proposed a novel self-supervised point cloud learning method by explicitly incorporating the local geometry information into the masked auto-encoding. Unlike popular methods that reconstructed masked cloud points from the unmasked cloud points, we validated that predicting the masked surfels is more effective, which was justified on six downstream tasks under various fine-tuning strategies. 
	Our method revealed the importance of local geometry in self-supervised point cloud learning, which could facilitate more subsequent studies in point cloud understanding.

	\appendix
	
	\section{Supplementary Material}
	
	\subsection{Task Settings on ScanObjectNN Dataset}
	We validate our MaskSurf with three task settings (\ie, OBJ-ONLY, OBJ-BG, and PB-T50-RS) on the ScanObjectNN dataset \cite{uy2019revisiting}.
	Specifically, the samples are segmented objects in the OBJ-ONLY setting, which is used to investigate the model robustness to deformed geometric shape and non-uniform surface density. 
	In the OBJ-BG setting, background points near the objects are also included, which is used to investigate the influence of background elements. 
	Additionally, to simulate more challenging cases in practice, bounding box perturbation is introduced. In the PB-T50-RS setting, the bounding boxes are randomly shifted up to 50\% of its size from the box centroid, and then rotated and scaled. 
	The PB-T50-RS setting is the most challenge one among all three settings.  
	
	\subsection{Domain Generalization on PointDA-10 Dataset}
	We investigate the synthetic-to-real domain generalization performance on the PointDA-10 dataset \cite{qin2019pointdan}, which includes two synthetic datasets of ModelNet-10 and ShapeNet-10 and one real-world dataset of ScanNet-10. 
	Specifically, samples of ModelNet-10, ShapeNet-10 and ScanNet-10 are from shared categories of ModelNet40 \cite{wu20153d}, ShapeNet \cite{shapenet2015}, and ScanNet \cite{dai2017scannet}, respectively.
	
	\subsection{Few-Shot Performance on ModelNet40}
	As illustrated in Tab. \ref{Tab:few_shot_modelnet40}, our MaskSurf consistently outperforms Point-MAE, justifying the advantage of masked surfel prediction over masked point prediction.
	
	\begin{table*} 
		\centering
		\caption{Few-shot classification performance on ModelNet40.} \label{Tab:few_shot_modelnet40}
		\begin{tabular}{lcccc}
			\hline
			& \multicolumn{2}{c}{\textbf{5-way}} & \multicolumn{2}{c}{\textbf{10-way}} \\
			\cmidrule(r){2-3} 	\cmidrule(r){4-5}
			& 10-shot & 20-shot & 10-shot & 20-shot \\
			\hline
			DGCNN \cite{wang2021unsupervised} & 31.6$\pm$2.8 & 40.8$\pm$4.6 & 19.9$\pm$2.1 & 16.9$\pm$1.5 \\
			Transformer \cite{vaswani2017attention} & 87.8$\pm$5.2 & 93.3$\pm$4.3 & 84.6$\pm$5.5 & 89.4$\pm$6.3 \\
			\hline
			\hline
			\multicolumn{5}{c}{\textbf{Transferring features protocol}} \\
			\hline
			DGCNN-OcCo \cite{wang2021unsupervised} & 90.6$\pm$2.8 & 92.5$\pm$1.9 & 82.9$\pm$1.3 & 86.5$\pm$2.2 \\
			Transformer-OcCo \cite{yu2021point} & 94.0$\pm$3.6 & 95.9$\pm$2.3 & 89.4$\pm$5.1 & 92.4$\pm$4.6 \\
			Point-BERT \cite{yu2021point} & 94.6$\pm$3.1 & 96.3$\pm$2.7 & 91.0$\pm$5.4 & 92.7$\pm$5.1 \\
			Point-MAE \cite{pang2022masked} & 96.3$\pm$2.5 & 97.8$\pm$1.8 & 92.6$\pm$4.1 & 95.0$\pm$3.0 \\
			MaskSurf (Ours) & \textbf{96.5}$\pm$2.5 & \textbf{98.0}$\pm$1.4 & \textbf{93.0}$\pm$4.1 & \textbf{95.3}$\pm$3.0   \\
			\hline
			\hline
			\multicolumn{5}{c}{\textbf{Linear classification protocol}}\\
			\hline
			Point-MAE \cite{pang2022masked} & 82.3$\pm$6.3 & 90.6$\pm$5.6 & 88.3$\pm$6.5 & \textbf{94.9}$\pm$3.5  \\
			MaskSurf (Ours) & \textbf{87.1}$\pm$4.6 & \textbf{92.3}$\pm$4.9 & \textbf{89.3}$\pm$4.2 & \textbf{94.9}$\pm$3.2 \\
			\hline
			\hline
			\multicolumn{5}{c}{\textbf{Non-linear classification protocol}}\\
			\hline
			Point-MAE \cite{pang2022masked} & 93.7$\pm$3.5 & 97.4$\pm$1.7 & \textbf{90.9}$\pm$5.0 & 94.2$\pm$4.2 \\
			MaskSurf (Ours) &  \textbf{95.4}$\pm$2.9 & \textbf{97.6}$\pm$1.4 & \textbf{90.9}$\pm$4.6 & \textbf{94.7}$\pm$3.3   \\
			\hline
		\end{tabular}
	\end{table*}
	
	\begin{table*} \footnotesize
		\centering
		\caption{Part segmentation results on the ShapeNetPart dataset. The mean IoU across all categories, \ie, mIoU$_{c}$ (\%), the mean IoU across all instances, \ie, mIoU$_{I}$ (\%), and IoU (\%) for each category are reported. } \label{Tab:partseg_category}
		\begin{tabular}{L{21.4mm}|C{5.0mm}C{5.7mm}|C{3.8mm}C{3.8mm}C{3.8mm}C{3.8mm}C{3.8mm}C{3.8mm}C{4.2mm}C{4.2mm}C{4.2mm}C{4.2mm}C{4.2mm}C{4.7mm}C{4.7mm}C{4.7mm}C{4.7mm}C{4.7mm}}
			\hline
			Methods & mIoU$_{c}$ & mIoU$_{I}$ & aero & bag & cap & car & chair & earph. & guitar & knife & lamp & laptop & motor & mug & pistol & rocket & skateb. & table \\
			\hline
			PointNet \cite{qi2017pointnet} & 80.39 & 83.7 & 83.4 & 78.7 & 82.5 & 74.9 & 89.6 & 73.0 & 91.5 & 85.9 & 80.8 & 95.3 & 65.2 & 93.0 & 81.2 & 57.9 & 72.8 & 80.6 \\
			PointNet++ \cite{qi2017pointnet++} & 81.85 & 85.1 & 82.4 & 79.0 & 87.7 & 77.3 & 90.8 & 71.8 & 91.0 & 85.9 & 83.7 & 95.3 & 71.6 & 94.1 & 81.3 & 58.7 & 76.4 & 82.6 \\
			DGCNN \cite{wang2019dynamic} & 82.33 & 85.2 & 84.0 & 83.4 & 86.7 & 77.8 & 90.6 & 74.7 & 91.2 & 87.5 & 82.8 & 95.7 & 66.3 & 94.9 & 81.1 & 63.5 & 74.5 & 82.6 \\
			Transformer \cite{vaswani2017attention} & 83.42 & 85.1 & 82.9 & 85.4 & 87.7 & 78.8 & 90.5 & 80.8 & 91.1 & 87.7 & 85.3 & 95.6 & 73.9 & 94.9 & 83.5 & 61.2 & 74.9 & 80.6 \\
			\hline
			\hline
			\multicolumn{18}{c}{\textbf{Transferring features protocol}}\\
			\hline
			Transformer-OcCo \cite{yu2021point} & 83.42 & 85.1 & 83.3 & 85.2 & 88.3 & 79.9 & 90.7 & 74.1 & 91.9 & 87.6 & 84.7 & 95.4 & 75.5 & 94.4 & 84.1 & 63.1 & 75.7 & 80.8   \\
			Point-BERT \cite{yu2021point} & 84.11 & 85.6 & 84.3 & 84.8 & 88.0 & 79.8 & 91.0 & 81.7 & 91.6 & 87.9 & 85.2 & 95.6 & 75.6 & 94.7 & 84.3 & 63.4 & 76.3 & 81.5  \\
			Point-MAE \cite{pang2022masked} & 84.19 & \textbf{86.1} & 84.3 & \textbf{85.0} & 88.3 & 80.5 & 91.3 & \textbf{78.5} & \textbf{92.1} & 87.4 & 86.1 & 96.1 & 75.2 & 94.6 & 84.7 & 63.5 & \textbf{77.1} & \textbf{82.4} \\
			MaskSurf (Ours) & \textbf{84.36} & \textbf{86.1} & \textbf{84.7} & 84.6 & \textbf{89.1} & \textbf{81.1} & \textbf{91.4} & 77.8 & 91.8 & \textbf{87.7} & \textbf{86.1} & \textbf{96.5} & \textbf{75.9} & \textbf{95.2} & \textbf{84.9} & \textbf{65.6} & 75.4 & 82.1 \\
			\hline
			\hline
			\multicolumn{18}{c}{\textbf{Non-linear classification protocol}} \\
			\hline
			Point-MAE \cite{pang2022masked} & 83.13 & 84.6 & 83.6 & 82.7 & 86.6 & 78.6 & 90.6 & 77.2 & 91.5 & 86.4 & 85.4 & 96.0 & 73.5 & 94.4 & 83.4 & 64.2 & 75.5 & 79.4  \\
			MaskSurf (Ours) & \textbf{83.30} & \textbf{85.3} & 82.9 & \textbf{82.9} & \textbf{87.4} & \textbf{79.0} & \textbf{90.7} & 72.0 & 91.3 & \textbf{86.5} & \textbf{85.8} & 95.7 & \textbf{74.6} & 94.1 & \textbf{83.7} & 62.1 & \textbf{76.3} & \textbf{81.2} \\
			\hline
		\end{tabular}
	\end{table*}
	
	\subsection{Classifier Architecture for Segmentation}
	We strictly follow \cite{pang2022masked} to construct the classifier for segmentation. Specifically, given learned features form the $4$th, $8$th and $12$th layers of Transformer block, we concatenate the multi-scale patch features and apply the max pooling and average pooling to them, resulting in two global feature representations. We follow \cite{qi2017pointnet++} to up-sample the concatenated path features to obtain interpolated features of each point.  
	In semantic segmentation, we concatenate the interpolated point features and two global features as complete point features. While in part segmentation, where the part label is associated to the object label, the complete point features are achieved by concatenating interpolated point features, two global features and one additional object feature, which are encoded with one FC layer from the object label.
	Finally, the point-wise prediction is obtained by forwarding the complete point features to three FC layers.

	\subsection{Segmentation Results and Visualizations}
\begin{figure*}[h]
		\vspace{-0.2cm}
		\begin{center}
			\includegraphics[width=0.82\linewidth]{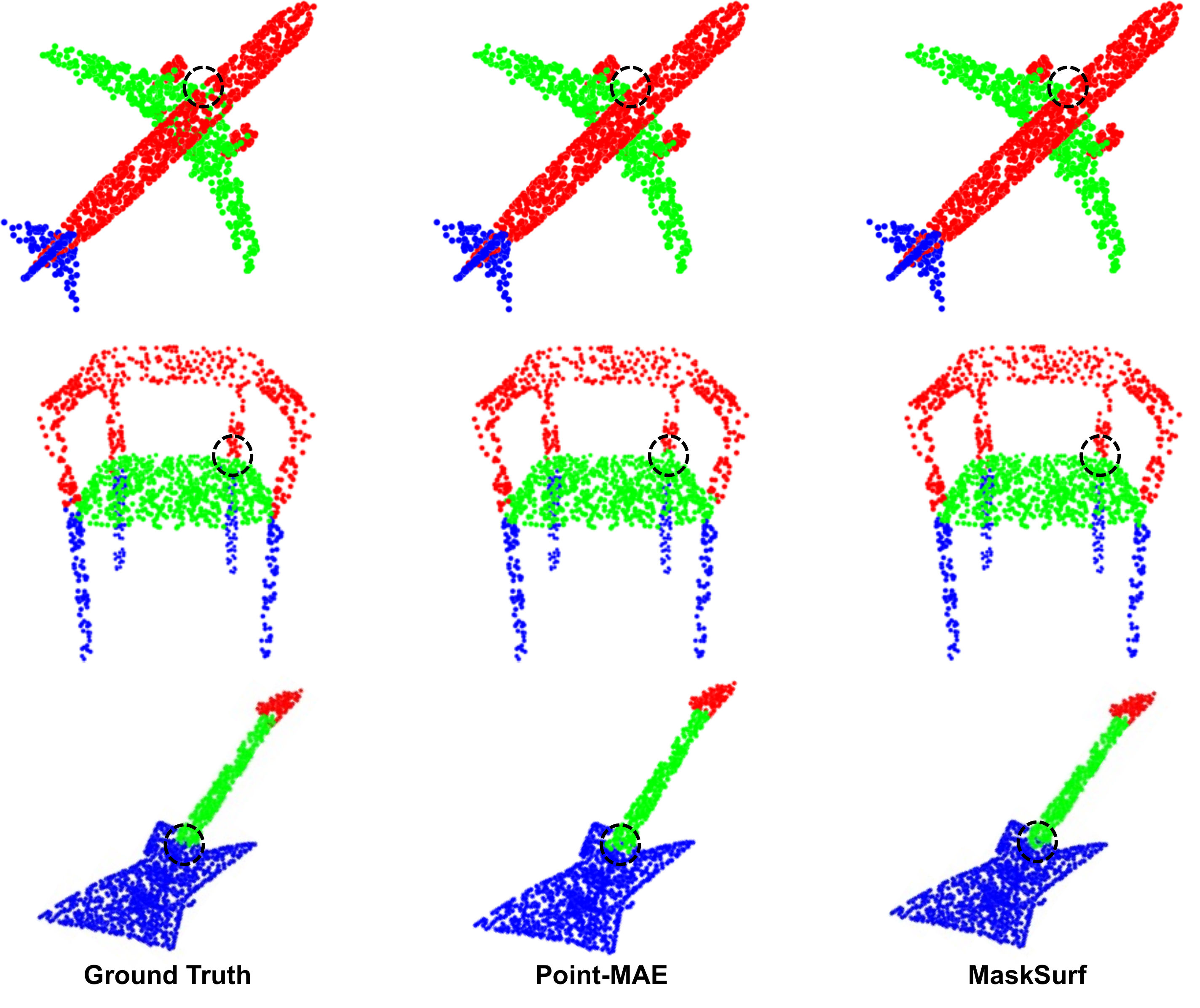}
		\end{center}
		\vspace{-0.2cm}
		\caption{Visualization of the part segmentation results on the ShapeNetPart test set.
		} \label{Fig:partseg_vis}
	\end{figure*}

\begin{figure*}[h]
		\vspace{-0.2cm}
		\begin{center}
			\includegraphics[width=0.95\linewidth]{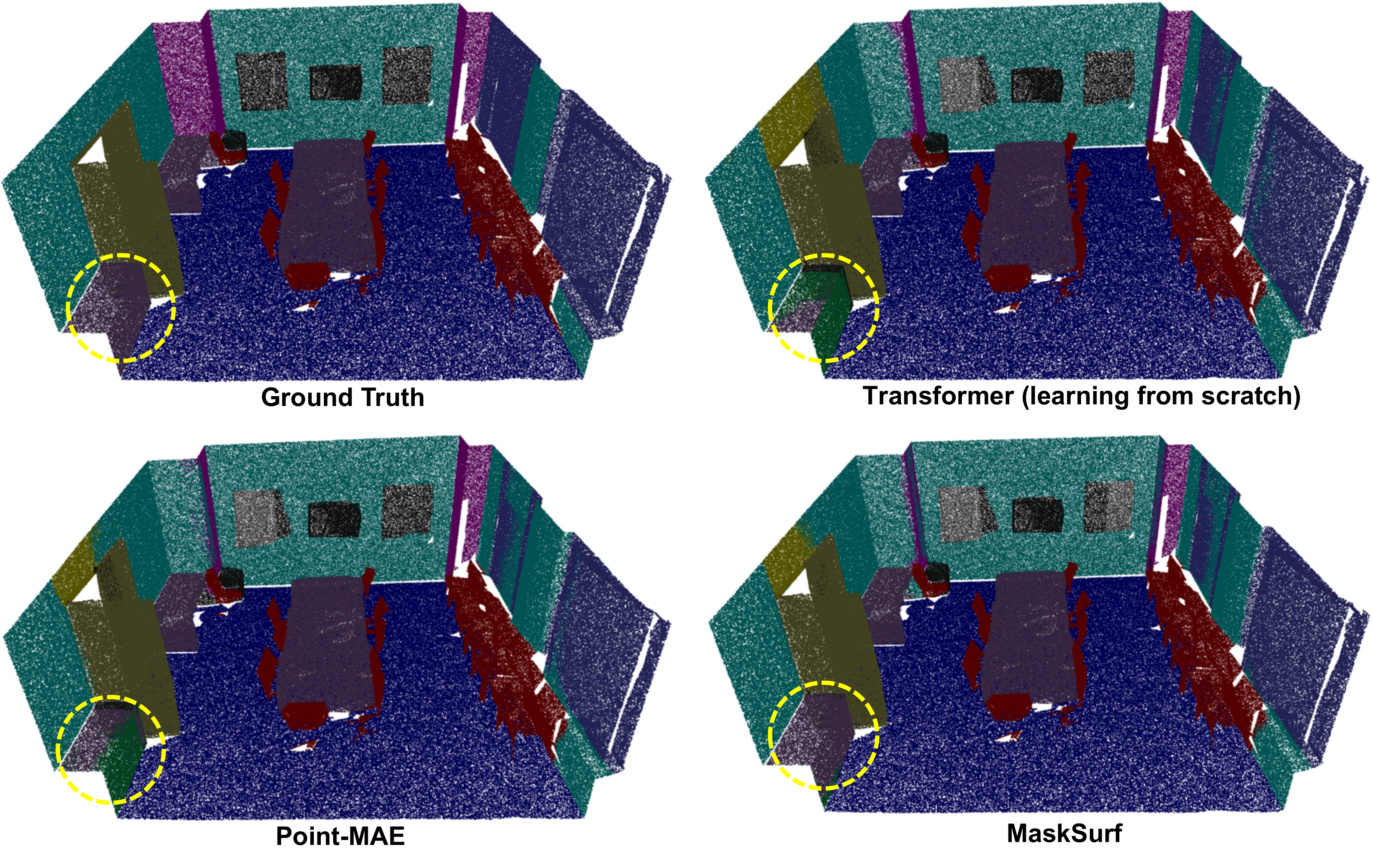}
		\end{center}
		\vspace{-0.2cm}
		\caption{Visualization of the semantic segmentation results on the S3DIS Area5.
		} \label{Fig:semseg_vis}
	\end{figure*}

	The detailed part segmentation results with category-wise mIoU are illustrated in Tab. \ref{Tab:partseg_category}.
	In addition, the results of part segmentation and semantic segmentation are visualized in Fig. \ref{Fig:partseg_vis} and Fig. \ref{Fig:semseg_vis}, respectively.

	\clearpage
	
	{\small
		\bibliographystyle{ieee_fullname}
		\bibliography{egbib}

\begin{thebibliography}{10}\itemsep=-1pt

\bibitem{afham2022crosspoint}
Mohamed Afham, Isuru Dissanayake, Dinithi Dissanayake, Amaya Dharmasiri,
  Kanchana Thilakarathna, and Ranga Rodrigo.
\newblock Crosspoint: Self-supervised cross-modal contrastive learning for 3d
  point cloud understanding.
\newblock {\em arXiv preprint arXiv:2203.00680}, 2022.

\bibitem{alexa2003computing}
Marc Alexa, Johannes Behr, Daniel Cohen-Or, Shachar Fleishman, David Levin, and
  Claudio~T. Silva.
\newblock Computing and rendering point set surfaces.
\newblock {\em IEEE Transactions on visualization and computer graphics},
  9(1):3--15, 2003.

\bibitem{armeni20163d}
Iro Armeni, Ozan Sener, Amir~R Zamir, Helen Jiang, Ioannis Brilakis, Martin
  Fischer, and Silvio Savarese.
\newblock 3d semantic parsing of large-scale indoor spaces.
\newblock In {\em Proceedings of the IEEE conference on computer vision and
  pattern recognition}, pages 1534--1543, 2016.

\bibitem{bae2008method}
Kwang-Ho Bae and Derek~D Lichti.
\newblock A method for automated registration of unorganised point clouds.
\newblock {\em ISPRS Journal of Photogrammetry and Remote Sensing},
  63(1):36--54, 2008.

\bibitem{shapenet2015}
Angel~X. Chang, Thomas Funkhouser, Leonidas Guibas, Pat Hanrahan, Qixing Huang,
  Zimo Li, Silvio Savarese, Manolis Savva, Shuran Song, Hao Su, Jianxiong Xiao,
  Li Yi, and Fisher Yu.
\newblock {ShapeNet: An Information-Rich 3D Model Repository}.
\newblock Technical Report arXiv:1512.03012 [cs.GR], Stanford University ---
  Princeton University --- Toyota Technological Institute at Chicago, 2015.

\bibitem{chen2020simple}
Ting Chen, Simon Kornblith, Mohammad Norouzi, and Geoffrey Hinton.
\newblock A simple framework for contrastive learning of visual
  representations.
\newblock In {\em International conference on machine learning}, pages
  1597--1607. PMLR, 2020.

\bibitem{chen2020improved}
Xinlei Chen, Haoqi Fan, Ross Girshick, and Kaiming He.
\newblock Improved baselines with momentum contrastive learning.
\newblock {\em arXiv preprint arXiv:2003.04297}, 2020.

\bibitem{chen2021shape}
Ye Chen, Jinxian Liu, Bingbing Ni, Hang Wang, Jiancheng Yang, Ning Liu, Teng
  Li, and Qi Tian.
\newblock Shape self-correction for unsupervised point cloud understanding.
\newblock In {\em Proceedings of the IEEE/CVF International Conference on
  Computer Vision}, pages 8382--8391, 2021.

\bibitem{cheng2021net}
Silin Cheng, Xiwu Chen, Xinwei He, Zhe Liu, and Xiang Bai.
\newblock Pra-net: Point relation-aware network for 3d point cloud analysis.
\newblock {\em IEEE Transactions on Image Processing}, 30:4436--4448, 2021.

\bibitem{dai2017scannet}
Angela Dai, Angel~X Chang, Manolis Savva, Maciej Halber, Thomas Funkhouser, and
  Matthias Nie{\ss}ner.
\newblock Scannet: Richly-annotated 3d reconstructions of indoor scenes.
\newblock In {\em Proceedings of the IEEE conference on computer vision and
  pattern recognition}, pages 5828--5839, 2017.

\bibitem{devlin2018bert}
Jacob Devlin, Ming-Wei Chang, Kenton Lee, and Kristina Toutanova.
\newblock Bert: Pre-training of deep bidirectional transformers for language
  understanding.
\newblock {\em arXiv preprint arXiv:1810.04805}, 2018.

\bibitem{dong2015image}
Chao Dong, Chen~Change Loy, Kaiming He, and Xiaoou Tang.
\newblock Image super-resolution using deep convolutional networks.
\newblock {\em IEEE transactions on pattern analysis and machine intelligence},
  38(2):295--307, 2015.

\bibitem{du2021self}
Bi'an Du, Xiang Gao, Wei Hu, and Xin Li.
\newblock Self-contrastive learning with hard negative sampling for
  self-supervised point cloud learning.
\newblock In {\em Proceedings of the 29th ACM International Conference on
  Multimedia}, pages 3133--3142, 2021.

\bibitem{fan2017point}
Haoqiang Fan, Hao Su, and Leonidas~J Guibas.
\newblock A point set generation network for 3d object reconstruction from a
  single image.
\newblock In {\em Proceedings of the IEEE conference on computer vision and
  pattern recognition}, pages 605--613, 2017.

\bibitem{fu2022pos}
Kexue Fu, Peng Gao, ShaoLei Liu, Renrui Zhang, Yu Qiao, and Manning Wang.
\newblock Pos-bert: Point cloud one-stage bert pre-training.
\newblock {\em arXiv preprint arXiv:2204.00989}, 2022.

\bibitem{gadelha2018multiresolution}
Matheus Gadelha, Rui Wang, and Subhransu Maji.
\newblock Multiresolution tree networks for 3d point cloud processing.
\newblock In {\em Proceedings of the European Conference on Computer Vision
  (ECCV)}, pages 103--118, 2018.

\bibitem{ganin2016domain}
Yaroslav Ganin, Evgeniya Ustinova, Hana Ajakan, Pascal Germain, Hugo
  Larochelle, Fran{\c{c}}ois Laviolette, Mario Marchand, and Victor Lempitsky.
\newblock Domain-adversarial training of neural networks.
\newblock {\em The journal of machine learning research}, 17(1):2096--2030,
  2016.

\bibitem{gidaris2018unsupervised}
Spyros Gidaris, Praveer Singh, and Nikos Komodakis.
\newblock Unsupervised representation learning by predicting image rotations.
\newblock {\em arXiv preprint arXiv:1803.07728}, 2018.

\bibitem{girshick2015fast}
Ross Girshick.
\newblock Fast r-cnn.
\newblock In {\em Proceedings of the IEEE international conference on computer
  vision}, pages 1440--1448, 2015.

\bibitem{goyal2021revisiting}
Ankit Goyal, Hei Law, Bowei Liu, Alejandro Newell, and Jia Deng.
\newblock Revisiting point cloud shape classification with a simple and
  effective baseline.
\newblock In {\em International Conference on Machine Learning}, pages
  3809--3820. PMLR, 2021.

\bibitem{grill2020bootstrap}
Jean-Bastien Grill, Florian Strub, Florent Altch{\'e}, Corentin Tallec, Pierre
  Richemond, Elena Buchatskaya, Carl Doersch, Bernardo Avila~Pires, Zhaohan
  Guo, Mohammad Gheshlaghi~Azar, et~al.
\newblock Bootstrap your own latent-a new approach to self-supervised learning.
\newblock {\em Advances in neural information processing systems},
  33:21271--21284, 2020.

\bibitem{gulrajani2020search}
Ishaan Gulrajani and David Lopez-Paz.
\newblock In search of lost domain generalization.
\newblock {\em arXiv preprint arXiv:2007.01434}, 2020.

\bibitem{guo2021pct}
Meng-Hao Guo, Jun-Xiong Cai, Zheng-Ning Liu, Tai-Jiang Mu, Ralph~R Martin, and
  Shi-Min Hu.
\newblock Pct: Point cloud transformer.
\newblock {\em Computational Visual Media}, 7(2):187--199, 2021.

\bibitem{habbecke2007surface}
Martin Habbecke and Leif Kobbelt.
\newblock A surface-growing approach to multi-view stereo reconstruction.
\newblock In {\em 2007 IEEE Conference on Computer Vision and Pattern
  Recognition}, pages 1--8. IEEE, 2007.

\bibitem{han2019multi}
Zhizhong Han, Xiyang Wang, Yu-Shen Liu, and Matthias Zwicker.
\newblock Multi-angle point cloud-vae: Unsupervised feature learning for 3d
  point clouds from multiple angles by joint self-reconstruction and
  half-to-half prediction.
\newblock In {\em 2019 IEEE/CVF International Conference on Computer Vision
  (ICCV)}, pages 10441--10450. IEEE, 2019.

\bibitem{hassani2019unsupervised}
Kaveh Hassani and Mike Haley.
\newblock Unsupervised multi-task feature learning on point clouds.
\newblock In {\em Proceedings of the IEEE/CVF International Conference on
  Computer Vision}, pages 8160--8171, 2019.

\bibitem{he2021masked}
Kaiming He, Xinlei Chen, Saining Xie, Yanghao Li, Piotr Doll{\'a}r, and Ross
  Girshick.
\newblock Masked autoencoders are scalable vision learners.
\newblock {\em arXiv preprint arXiv:2111.06377}, 2021.

\bibitem{he2020momentum}
Kaiming He, Haoqi Fan, Yuxin Wu, Saining Xie, and Ross Girshick.
\newblock Momentum contrast for unsupervised visual representation learning.
\newblock In {\em Proceedings of the IEEE/CVF conference on computer vision and
  pattern recognition}, pages 9729--9738, 2020.

\bibitem{he2017mask}
Kaiming He, Georgia Gkioxari, Piotr Doll{\'a}r, and Ross Girshick.
\newblock Mask r-cnn.
\newblock In {\em Proceedings of the IEEE international conference on computer
  vision}, pages 2961--2969, 2017.

\bibitem{he2016deep}
Kaiming He, Xiangyu Zhang, Shaoqing Ren, and Jian Sun.
\newblock Deep residual learning for image recognition.
\newblock In {\em Proceedings of the IEEE conference on computer vision and
  pattern recognition}, pages 770--778, 2016.

\bibitem{huang2021spatio}
Siyuan Huang, Yichen Xie, Song-Chun Zhu, and Yixin Zhu.
\newblock Spatio-temporal self-supervised representation learning for 3d point
  clouds.
\newblock In {\em Proceedings of the IEEE/CVF International Conference on
  Computer Vision}, pages 6535--6545, 2021.

\bibitem{krizhevsky2012imagenet}
Alex Krizhevsky, Ilya Sutskever, and Geoffrey~E Hinton.
\newblock Imagenet classification with deep convolutional neural networks.
\newblock {\em Advances in neural information processing systems}, 25, 2012.

\bibitem{li2018pointcnn}
Yangyan Li, Rui Bu, Mingchao Sun, Wei Wu, Xinhan Di, and Baoquan Chen.
\newblock Pointcnn: Convolution on x-transformed points.
\newblock {\em Advances in neural information processing systems}, 31, 2018.

\bibitem{liu2022masked}
Haotian Liu, Mu Cai, and Yong~Jae Lee.
\newblock Masked discrimination for self-supervised learning on point clouds.
\newblock {\em arXiv preprint arXiv:2203.11183}, 2022.

\bibitem{liu2019relation}
Yongcheng Liu, Bin Fan, Shiming Xiang, and Chunhong Pan.
\newblock Relation-shape convolutional neural network for point cloud analysis.
\newblock In {\em Proceedings of the IEEE/CVF Conference on Computer Vision and
  Pattern Recognition}, pages 8895--8904, 2019.

\bibitem{loshchilov2016sgdr}
Ilya Loshchilov and Frank Hutter.
\newblock Sgdr: Stochastic gradient descent with warm restarts.
\newblock {\em arXiv preprint arXiv:1608.03983}, 2016.

\bibitem{loshchilov2017decoupled}
Ilya Loshchilov and Frank Hutter.
\newblock Decoupled weight decay regularization.
\newblock {\em arXiv preprint arXiv:1711.05101}, 2017.

\bibitem{ma2022rethinking}
Xu Ma, Can Qin, Haoxuan You, Haoxi Ran, and Yun Fu.
\newblock Rethinking network design and local geometry in point cloud: A simple
  residual mlp framework.
\newblock {\em arXiv preprint arXiv:2202.07123}, 2022.

\bibitem{noroozi2016unsupervised}
Mehdi Noroozi and Paolo Favaro.
\newblock Unsupervised learning of visual representations by solving jigsaw
  puzzles.
\newblock In {\em European conference on computer vision}, pages 69--84.
  Springer, 2016.

\bibitem{pang2022masked}
Yatian Pang, Wenxiao Wang, Francis~EH Tay, Wei Liu, Yonghong Tian, and Li Yuan.
\newblock Masked autoencoders for point cloud self-supervised learning.
\newblock {\em arXiv preprint arXiv:2203.06604}, 2022.

\bibitem{pauly2003shape}
Mark Pauly, Richard Keiser, Leif~P Kobbelt, and Markus Gross.
\newblock Shape modeling with point-sampled geometry.
\newblock {\em ACM Transactions on Graphics (TOG)}, 22(3):641--650, 2003.

\bibitem{pfister2000surfels}
Hanspeter Pfister, Matthias Zwicker, Jeroen Van~Baar, and Markus Gross.
\newblock Surfels: Surface elements as rendering primitives.
\newblock In {\em Proceedings of the 27th annual conference on Computer
  graphics and interactive techniques}, pages 335--342, 2000.

\bibitem{poursaeed2020self}
Omid Poursaeed, Tianxing Jiang, Han Qiao, Nayun Xu, and Vladimir~G Kim.
\newblock Self-supervised learning of point clouds via orientation estimation.
\newblock In {\em 2020 International Conference on 3D Vision (3DV)}, pages
  1018--1028. IEEE, 2020.

\bibitem{qi2017pointnet}
Charles~R Qi, Hao Su, Kaichun Mo, and Leonidas~J Guibas.
\newblock Pointnet: Deep learning on point sets for 3d classification and
  segmentation.
\newblock In {\em Proceedings of the IEEE conference on computer vision and
  pattern recognition}, pages 652--660, 2017.

\bibitem{qi2017pointnet++}
Charles~Ruizhongtai Qi, Li Yi, Hao Su, and Leonidas~J Guibas.
\newblock Pointnet++: Deep hierarchical feature learning on point sets in a
  metric space.
\newblock {\em Advances in neural information processing systems}, 30, 2017.

\bibitem{qin2019pointdan}
Can Qin, Haoxuan You, Lichen Wang, C-C~Jay Kuo, and Yun Fu.
\newblock Pointdan: A multi-scale 3d domain adaption network for point cloud
  representation.
\newblock {\em Advances in Neural Information Processing Systems}, 32, 2019.

\bibitem{qiu2021geometric}
Shi Qiu, Saeed Anwar, and Nick Barnes.
\newblock Geometric back-projection network for point cloud classification.
\newblock {\em IEEE Transactions on Multimedia}, 2021.

\bibitem{ran2022surface}
Haoxi Ran, Jun Liu, and Chengjie Wang.
\newblock Surface representation for point clouds.
\newblock In {\em Proceedings of the IEEE/CVF Conference on Computer Vision and
  Pattern Recognition}, pages 18942--18952, 2022.

\bibitem{rao2020global}
Yongming Rao, Jiwen Lu, and Jie Zhou.
\newblock Global-local bidirectional reasoning for unsupervised representation
  learning of 3d point clouds.
\newblock In {\em Proceedings of the IEEE/CVF Conference on Computer Vision and
  Pattern Recognition}, pages 5376--5385, 2020.

\bibitem{ronneberger2015u}
Olaf Ronneberger, Philipp Fischer, and Thomas Brox.
\newblock U-net: Convolutional networks for biomedical image segmentation.
\newblock In {\em International Conference on Medical image computing and
  computer-assisted intervention}, pages 234--241. Springer, 2015.

\bibitem{sanghi2020info3d}
Aditya Sanghi.
\newblock Info3d: Representation learning on 3d objects using mutual
  information maximization and contrastive learning.
\newblock In {\em European Conference on Computer Vision}, pages 626--642.
  Springer, 2020.

\bibitem{sauder2019self}
Jonathan Sauder and Bjarne Sievers.
\newblock Self-supervised deep learning on point clouds by reconstructing
  space.
\newblock {\em Advances in Neural Information Processing Systems}, 32, 2019.

\bibitem{lulu2020improving}
Lulu Tang, Ke Chen, Chaozheng Wu, Yu Hong, Kui Jia, and Zhi-Xin Yang.
\newblock Improving semantic analysis on point clouds via auxiliary supervision
  of local geometric priors.
\newblock {\em IEEE Transactions on Cybernetics}, pages 1--11, 2020.

\bibitem{tatarchenko2018tangent}
Maxim Tatarchenko, Jaesik Park, Vladlen Koltun, and Qian-Yi Zhou.
\newblock Tangent convolutions for dense prediction in 3d.
\newblock In {\em Proceedings of the IEEE Conference on Computer Vision and
  Pattern Recognition}, pages 3887--3896, 2018.

\bibitem{thomas2019kpconv}
Hugues Thomas, Charles~R Qi, Jean-Emmanuel Deschaud, Beatriz Marcotegui,
  Fran{\c{c}}ois Goulette, and Leonidas~J Guibas.
\newblock Kpconv: Flexible and deformable convolution for point clouds.
\newblock In {\em Proceedings of the IEEE/CVF international conference on
  computer vision}, pages 6411--6420, 2019.

\bibitem{tian2019fcos}
Zhi Tian, Chunhua Shen, Hao Chen, and Tong He.
\newblock Fcos: Fully convolutional one-stage object detection.
\newblock In {\em Proceedings of the IEEE/CVF international conference on
  computer vision}, pages 9627--9636, 2019.

\bibitem{tong2022videomae}
Zhan Tong, Yibing Song, Jue Wang, and Limin Wang.
\newblock Videomae: Masked autoencoders are data-efficient learners for
  self-supervised video pre-training.
\newblock {\em arXiv preprint arXiv:2203.12602}, 2022.

\bibitem{uy2019revisiting}
Mikaela~Angelina Uy, Quang-Hieu Pham, Binh-Son Hua, Thanh Nguyen, and Sai-Kit
  Yeung.
\newblock Revisiting point cloud classification: A new benchmark dataset and
  classification model on real-world data.
\newblock In {\em Proceedings of the IEEE/CVF international conference on
  computer vision}, pages 1588--1597, 2019.

\bibitem{vaswani2017attention}
Ashish Vaswani, Noam Shazeer, Niki Parmar, Jakob Uszkoreit, Llion Jones,
  Aidan~N Gomez, {\L}ukasz Kaiser, and Illia Polosukhin.
\newblock Attention is all you need.
\newblock {\em Advances in neural information processing systems}, 30, 2017.

\bibitem{wang2021unsupervised}
Hanchen Wang, Qi Liu, Xiangyu Yue, Joan Lasenby, and Matt~J Kusner.
\newblock Unsupervised point cloud pre-training via occlusion completion.
\newblock In {\em Proceedings of the IEEE/CVF International Conference on
  Computer Vision}, pages 9782--9792, 2021.

\bibitem{wang2019dynamic}
Yue Wang, Yongbin Sun, Ziwei Liu, Sanjay~E Sarma, Michael~M Bronstein, and
  Justin~M Solomon.
\newblock Dynamic graph cnn for learning on point clouds.
\newblock {\em Acm Transactions On Graphics (tog)}, 38(5):1--12, 2019.

\bibitem{wei2021masked}
Chen Wei, Haoqi Fan, Saining Xie, Chao-Yuan Wu, Alan Yuille, and Christoph
  Feichtenhofer.
\newblock Masked feature prediction for self-supervised visual pre-training.
\newblock {\em arXiv preprint arXiv:2112.09133}, 2021.

\bibitem{weise2009hand}
Thibaut Weise, Thomas Wismer, Bastian Leibe, and Luc Van~Gool.
\newblock In-hand scanning with online loop closure.
\newblock In {\em 2009 IEEE 12th International Conference on Computer Vision
  Workshops, ICCV Workshops}, pages 1630--1637. IEEE, 2009.

\bibitem{wu20153d}
Zhirong Wu, Shuran Song, Aditya Khosla, Fisher Yu, Linguang Zhang, Xiaoou Tang,
  and Jianxiong Xiao.
\newblock 3d shapenets: A deep representation for volumetric shapes.
\newblock In {\em Proceedings of the IEEE conference on computer vision and
  pattern recognition}, pages 1912--1920, 2015.

\bibitem{xu2022cp}
Mingye Xu, Zhipeng Zhou, Hongbin Xu, Yali Wang, and Yu Qiao.
\newblock Cp-net: Contour-perturbed reconstruction network for self-supervised
  point cloud learning.
\newblock {\em arXiv preprint arXiv:2201.08215}, 2022.

\bibitem{xu2018spidercnn}
Yifan Xu, Tianqi Fan, Mingye Xu, Long Zeng, and Yu Qiao.
\newblock Spidercnn: Deep learning on point sets with parameterized
  convolutional filters.
\newblock In {\em Proceedings of the European Conference on Computer Vision
  (ECCV)}, pages 87--102, 2018.

\bibitem{yan2022implicit}
Siming Yan, Zhenpei Yang, Haoxiang Li, Li Guan, Hao Kang, Gang Hua, and Qixing
  Huang.
\newblock Implicit autoencoder for point cloud self-supervised representation
  learning.
\newblock {\em arXiv preprint arXiv:2201.00785}, 2022.

\bibitem{yang2018foldingnet}
Yaoqing Yang, Chen Feng, Yiru Shen, and Dong Tian.
\newblock Foldingnet: Point cloud auto-encoder via deep grid deformation.
\newblock In {\em Proceedings of the IEEE conference on computer vision and
  pattern recognition}, pages 206--215, 2018.

\bibitem{yi2016scalable}
Li Yi, Vladimir~G Kim, Duygu Ceylan, I-Chao Shen, Mengyan Yan, Hao Su, Cewu Lu,
  Qixing Huang, Alla Sheffer, and Leonidas Guibas.
\newblock A scalable active framework for region annotation in 3d shape
  collections.
\newblock {\em ACM Transactions on Graphics (ToG)}, 35(6):1--12, 2016.

\bibitem{yu2021point}
Xumin Yu, Lulu Tang, Yongming Rao, Tiejun Huang, Jie Zhou, and Jiwen Lu.
\newblock Point-bert: Pre-training 3d point cloud transformers with masked
  point modeling.
\newblock {\em arXiv preprint arXiv:2111.14819}, 2021.

\bibitem{zhang2021pvt}
Cheng Zhang, Haocheng Wan, Shengqiang Liu, Xinyi Shen, and Zizhao Wu.
\newblock Pvt: Point-voxel transformer for 3d deep learning.
\newblock {\em arXiv preprint arXiv:2108.06076}, 2021.

\bibitem{zhang2017beyond}
Kai Zhang, Wangmeng Zuo, Yunjin Chen, Deyu Meng, and Lei Zhang.
\newblock Beyond a gaussian denoiser: Residual learning of deep cnn for image
  denoising.
\newblock {\em IEEE transactions on image processing}, 26(7):3142--3155, 2017.

\bibitem{zhang2022point}
Renrui Zhang, Ziyu Guo, Peng Gao, Rongyao Fang, Bin Zhao, Dong Wang, Yu Qiao,
  and Hongsheng Li.
\newblock Point-m2ae: Multi-scale masked autoencoders for hierarchical point
  cloud pre-training.
\newblock {\em arXiv preprint arXiv:2205.14401}, 2022.

\bibitem{zhao2021point}
Hengshuang Zhao, Li Jiang, Jiaya Jia, Philip~HS Torr, and Vladlen Koltun.
\newblock Point transformer.
\newblock In {\em Proceedings of the IEEE/CVF International Conference on
  Computer Vision}, pages 16259--16268, 2021.

\bibitem{zhao20193d}
Yongheng Zhao, Tolga Birdal, Haowen Deng, and Federico Tombari.
\newblock 3d point capsule networks.
\newblock In {\em Proceedings of the IEEE/CVF Conference on Computer Vision and
  Pattern Recognition}, pages 1009--1018, 2019.

\bibitem{zhou2022self}
Junsheng Zhou, Xin Wen, Yu-Shen Liu, Yi Fang, and Zhizhong Han.
\newblock Self-supervised point cloud representation learning with occlusion
  auto-encoder.
\newblock {\em arXiv preprint arXiv:2203.14084}, 2022.

\end{thebibliography}
	}
	
\end{document}